\documentclass[preprint,6pt]{elsarticle}

\usepackage[utf8x]{inputenc}
\usepackage[title]{appendix}
\usepackage{amssymb}
\usepackage{graphics}
\usepackage{graphicx}
\usepackage{subfigure}
\usepackage{epsfig}
\usepackage{amsmath}
\usepackage{bm}
\usepackage{lineno}
\usepackage{url}
\usepackage{float}
\usepackage{color}
\usepackage[normalem]{ulem}
\usepackage[color,final]{showkeys}
\usepackage{multirow}

\usepackage{amssymb}
\usepackage{mathrsfs}
\usepackage{graphics}
\usepackage{graphicx}
\usepackage{subfigure}
\usepackage{epsfig}
\usepackage{epstopdf}
\usepackage{amsmath}
\usepackage{bm}
\usepackage{lineno}
\usepackage{url}
\usepackage{float}
\usepackage{color}
\usepackage{amssymb}
\usepackage{multirow}
\usepackage{subfigure}
\usepackage{graphicx}
\usepackage{amsmath}
\usepackage[normalem]{ulem}
\usepackage[color,final]{showkeys}
\usepackage{multirow}
\usepackage{setspace}

\usepackage{float}
\usepackage{marvosym}
\usepackage{multirow}
\usepackage{graphicx}
\usepackage{subfigure}
\usepackage{amsfonts}
\usepackage{amsmath}
\usepackage{amssymb}
\usepackage{algorithm}
\usepackage{algorithmic}
\usepackage[colorlinks,
linkcolor=blue,
anchorcolor=green,
urlcolor=blue,
citecolor=green]{hyperref}
\usepackage{tikz,xcolor,hyperref}

\usepackage{orcidlink}
\usepackage{array}
\usepackage{soul}
\usepackage{xcolor}
\usepackage{natbib}
\usepackage{utfsym}
\usepackage{amsmath,amsfonts,amssymb}
\usepackage{graphicx}
\usepackage{setspace}
\usepackage{booktabs}
\usepackage[subfigure]{tocloft}
\usepackage{amssymb}
\usepackage{mathrsfs}
\usepackage{subfigure}
\usepackage{epstopdf}
\usepackage{tocloft}
\usepackage{hyperref}
\usepackage[figure]{hypcap}
\usepackage{algorithmic}
\usepackage{algorithm}
\usepackage{comment}
\usepackage{rotating}
\soulregister\cite7
\soulregister\ref7
\soulregister\eqref7

\journal{J COMPUT APPL MATH}

\begin{document}

\begin{frontmatter}
%\title{A New Pixel Intensity Constraint Framework for Blind Binary and Pattern Image Deconvolution}
\title{Blind Deconvolution of Binary and Pattern Images with Pixel Intensity Constraints and Sparse Gradient Prior}

%\tnotetext[label1]{
%%This work was supported in part by the NSFC under Grants 12001005, 61806134 and 62076170; in part by the start-up funding of Anhui University under Grant Y040418173; in part by the Natural Science Research Project of Anhui Universities under Grant KJ2019A0032; in part by the Project of Natural Science Foundation of Anhui Province under Grant 2008085QF286;   in part by the Project of Science and Technology Planning Project of Sichuan Province under Grant 2020YFG0324;  in part by the Project of Sichuan University Innovation Spark Project under Grant 2019SCUH0007.
%*\textbf{Corresponding author.}}

\author[label1]{Qinghua Zhang\fnref{fn1}}
\ead{qhzhang@mail.bnu.edu.cn}

\author[label2]{Xuesong Yang\fnref{fn1}}
\ead{xuesongy@stu.ahu.edu.cn}

\author[label2]{Liangtian He\corref{cor1}}
\ead{helt@ahu.edu.cn}

\author[label3]{\\Liang-jian Deng}
\ead{liangjian.deng@uestc.edu.cn}

\author[label4]{Jun Liu}
\ead{liuj292@nenu.edu.cn}

\address[label1]{Laboratory of Mathematics and Complex Systems
(Ministry of Education of China), the School of Mathematical Sciences,
Beijing Normal University, Beijing 100875, P. R. China}

\address[label2]{School of Mathematical Sciences, and Anhui University
Center for Applied Mathematics, Anhui University, Hefei, 230601, P. R. China}

\address[label3]{School of Mathematical Sciences, University of Electronic Science and Technology of China, Chengdu 611731,  P. R. China}

\address[label4]{Key Laboratory for Applied Statistics of MOE, School of Mathematics
and Statistics, Northeast Normal University, Changchun 130024, P. R. China}

\fntext[fn1]{Qinghua Zhang and Xuesong Yang contributed equally to this work.}

\cortext[cor1]{Corresponding author: Liangtian He.}

\begin{abstract}
Blind image deconvolution (BID) is a prominent research topic in the field of imaging sciences, given its significant practical applications. 
Most existing model-based BID methods focus on natural images, incorporating appropriate prior knowledge about both the underlying image and the blur kernel. However, for certain classes of images, such as barcodes, text, and patterns, pixels can only take very limited values, a specific prior that is often overlooked in the literature.
In this article, we introduce a novel pixel intensity constraint to leverage this important information, improving recovery performance for these specialized image classes. Specifically, we propose a unified framework for blind binary and pattern image deconvolution that incorporates both the pixel intensity constraint and a gradient sparsity regularizer.
Numerical experiments demonstrate that our method outperforms many existing BID techniques, achieving superior results in terms of both visual quality and quantitative metrics. The data and code associated with this article are available at: {\color{blue}\url{https://github.com/xuesyang/PIC-BID}}.

\end{abstract}

\begin{keyword}
Blind deconvolution,  binary image,  pattern image,   pixel intensity constraint,  gradient sparsity.
\end{keyword}

\end{frontmatter}

\section{Introduction}
Due to the inevitable external factors and imperfections of imaging systems, the recorded images are often corrupted by blur as well as additive noise. For instance, image blurring often arises from optical aberrations, object motion during the exposure time, while noise can be caused by malfunctioning pixels in camera sensors. Generally speaking, when the blur is uniform and spatially invariant, the blur degradation can often be mathematically formulated by 
\begin{equation}
\mathbf{y}= \mathbf{k} \ast \mathbf{x} + \mathbf{n},
\end{equation}
The task of image deconvolution/deblurring (ID) is to recover the sharp image from a blurry measurement, which is primarily categorized into two groups: non-blind deconvolution and blind deconvolution. 
In scenarios where the blur kernel is known beforehand, this is the classic non-blind image deconvolution (NBID) problem \cite{xu2014domain, huang2019nonstationary}. In contrast, when the blur kernel is unknown, both the latent image and the blur kernel need to be estimated. Blind image deconvolution (BID) presents greater challenges than its NBID counterpart. Nevertheless, a variety of effective BID methods have been investigated over the past decades, which are broadly classified into two categories:  the model-driven optimization methods \cite{li2018regularized,fergus2006removing,shan2008high,cai2011framelet,lou2014partially,
van2015regularization,perrone2015clearer,pan2016blind,shao2020gradient,liu2021surface,
ge2021blind,hu2022salient,zhang2022pixel,ge2022blind,yan2017image} and data-driven deep learning methods \cite{schuler2015learning,nah2017deep,kupyn2018deblurgan,ren2020neural,pan2020physics,
tran2021explore,zamir2021multi,li2022learning,rim2022realistic,kim2022mssnet,
cui2023exploring,ferreira2023restoring,zhang2023self}.

It is worth mentioning that the BID problem is highly ill-posed because many different pairs of $\mathbf{x}$ and $\mathbf{k}$ can give rise to the same $\mathbf{y}$,  such as blurred images and \textit{delta} blur filters.
For model-based BID methods, which are commonly used techniques, several kinds of prior information about both the sharp image and the blur kernel need to be exploited to alleviate the ill-posed nature of BID.
This methodology aims to achieve robust and promising results by constraining the solution space, which can be viewed from the perspective of Bayesian estimation.
More specifically, in terms of maximum a posteriori probability (MAP) estimation, one seeks the latent sharp image $\mathbf{x}$ and the unknown blur kernel $\mathbf{k}$, given the corrupted observation $\mathbf{y}$, that maximize the posterior probability 
$P(\mathbf{x},\mathbf{k}|\mathbf{y})$ as follows. 
\begin{equation}
\begin{aligned}
(\mathbf{k},\mathbf{x})&=\mathop{\arg\max}\limits_{\mathbf{k},\mathbf{x}}P(\mathbf{k},\mathbf{x}|\mathbf{y})\\
&=\mathop{\arg\max}\limits_{\mathbf{k},\mathbf{x}}P(\mathbf{y}|\mathbf{k},\mathbf{x})P(\mathbf{x})P(\mathbf{k})\\
&=\mathop{\arg\min}\limits_{\mathbf{k},\mathbf{x}} - \mathrm{log} P(\mathbf{y}|\mathbf{k},\mathbf{x}) - \mathrm{log} P(\mathbf{x}) - \mathrm{log} P(\mathbf{k}),
\end{aligned}
\end{equation}
where the monotonic decreasing function $-\mathrm{log}(\cdot)$ is introduced here to recast the maximization as a simpler minimization problem. $P(\mathbf{y}|\mathbf{k},\mathbf{x})$ is the likelihood corresponding to the fidelity term, and $P(\mathbf{x})$ and $P(\mathbf{k})$ model the priors of the clean image and blur kernel, respectively.
Consequently, armed with a clear probability relation between the degraded observation $\mathbf{y}$,  latent image $\mathbf{x}$ and blur kernel $\mathbf{k}$, also with trusted priors,
the MAP derivation of (2) can be rewritten as the following minimization problem.
\begin{equation}
    \mathop{\arg\min}\limits_{\mathbf{k},\mathbf{x}}\frac{1}{2}\|\mathbf{k}* \mathbf{x}-\mathbf{y}\|_{F}^{2}+\lambda_{1}\mathcal{R}(\mathbf{x})+\lambda_{2}\mathcal{R}(\mathbf{k}),
\end{equation}
where $\mathcal{R}(\mathbf{x}), \mathcal{R}(\mathbf{k})$ represent the regularization terms associated with the sharp image and the blur kernel respectively, and $\lambda_{1},\lambda_{2}$ denotes the regularization parameters which balance these terms.
During the last several decades, there have been sustained efforts in exploring image priors, ranging from the sparsity-inspired prior \cite{shan2008high,cai2011framelet,perrone2015clearer} and sophisticated 
low-rank prior \cite{chen2018algorithm,chi2019nonconvex,yuan2023rank} to the dark channel 
prior \cite{pan2016blind,ge2021blind,chen2019blind,wen2020simple,liu2022blind}, just 
to mention a few.
The empirical success of these approaches demonstrates that MAP estimation is capable of delivering favorable BID results when appropriate priors are selected.
How to determine the two terms $\mathcal{R}(\mathbf{x})$ and $\mathcal{R}(\mathbf{k})$, this is the key component of model-driven BID methods.
Roughly speaking, it may be argued that almost everything done in this field revolves around the quest for finding reliable priors and formulating proper regularizer terms.

\begin{figure}[t]
  \centering
  \subfigure[]{\raisebox{0.5cm}{\includegraphics[scale=0.16]{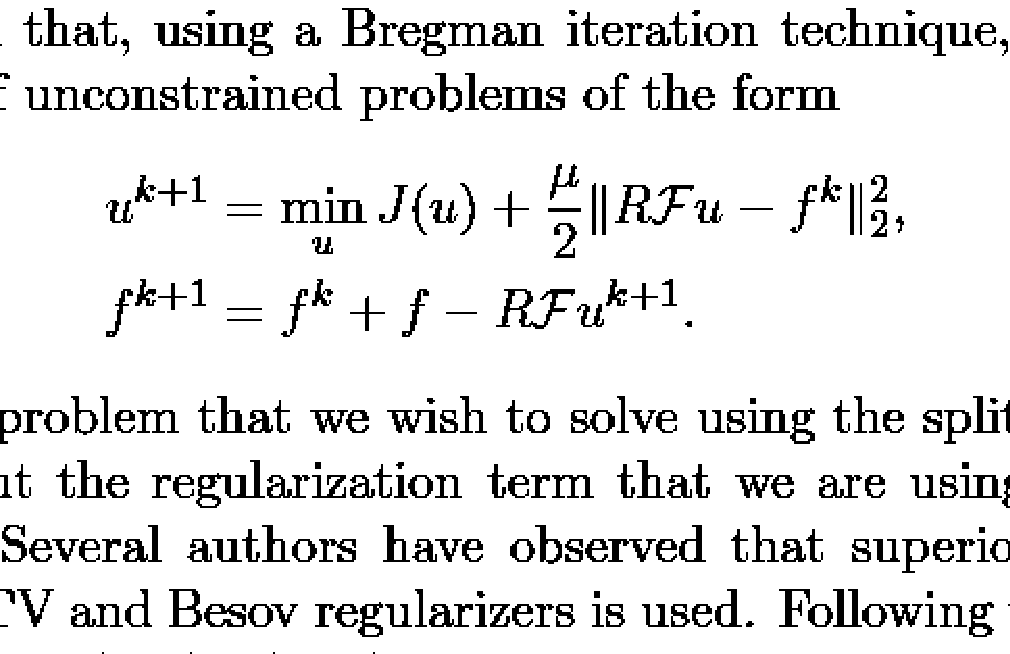}}}
  \subfigure[]{\includegraphics[scale=0.24]{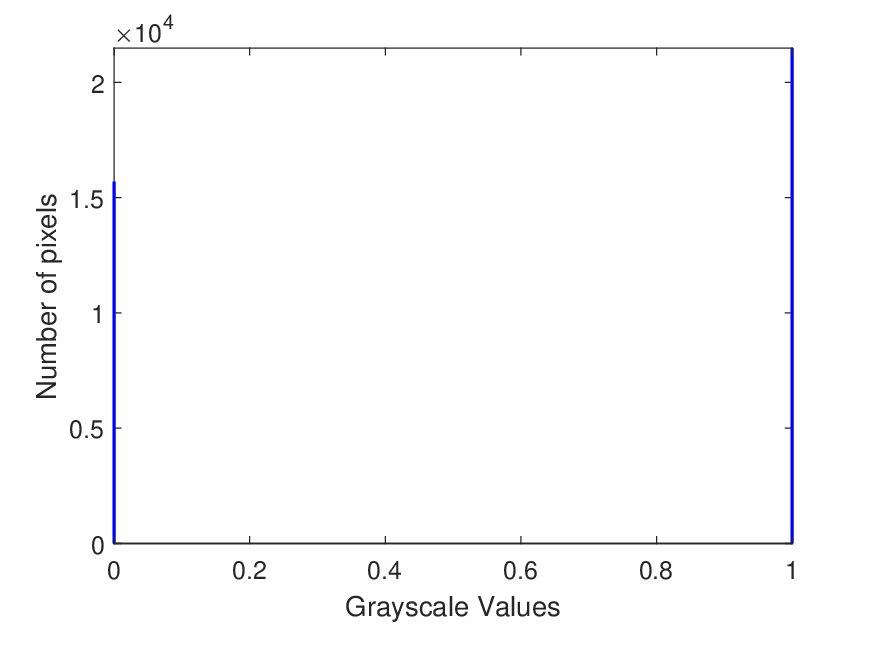}}
  \subfigure[]{\raisebox{0.5cm}{\includegraphics[scale=0.12]{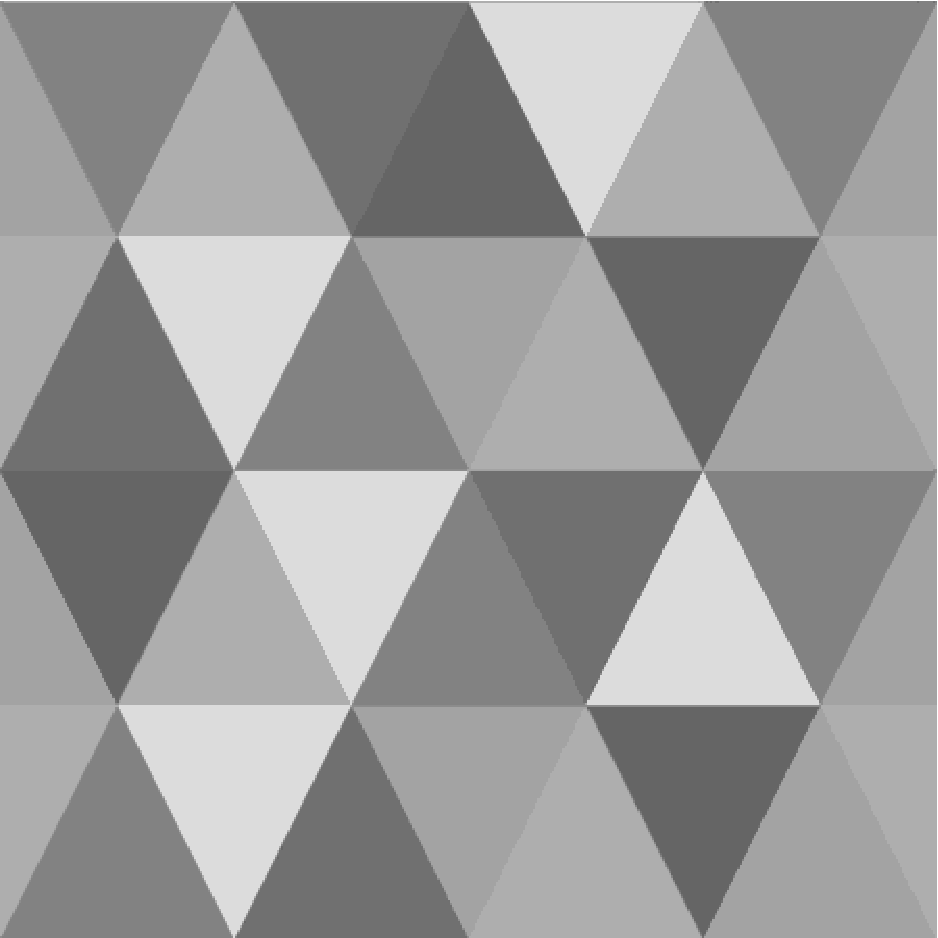}}}
  \subfigure[]{\includegraphics[scale=0.24]{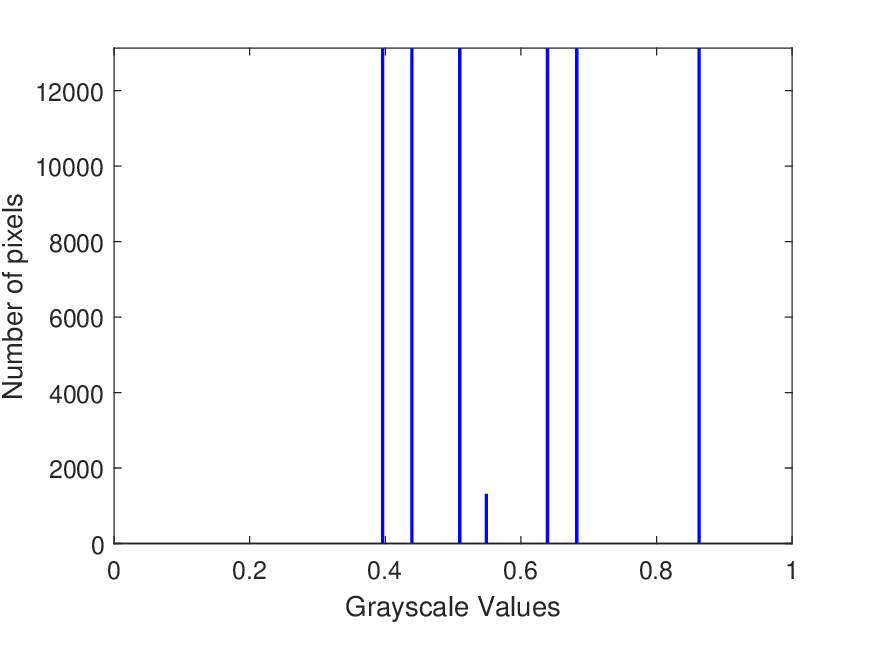}}

  \caption{An illustrative example of the two-valued \textit{binary} image and the multi-valued \textit{pattern} images. (a) Binary image and (b) its pixel intensity histogram. (c) Pattern image and (d) its pixel intensity histogram.}
\label{Fig:12}
\end{figure}

In the literature, regarding the process of the latent image and blur kernel, BID methods can be further classified into two categories \cite{liu2021surface}. The first category involves estimating the sharp image and blur kernel simultaneously \cite{shan2008high,cai2011framelet}. The second category involves a popular two-step strategy: first estimating the blur kernel, followed by applying a non-blind deconvolution approach to obtain a better result \cite{cho2011handling,whyte2014deblurring}. 
Notably, our algorithm adopts the first approach. As we will show in Section 4 with experimental results, simultaneously estimating the image and blur kernel in the proposed algorithm yields superior results for blind binary and pattern image deblurring compared to the second approach.

\begin{figure}
  \centering
  \subfigure[Ground truth]{\includegraphics[scale=0.23]{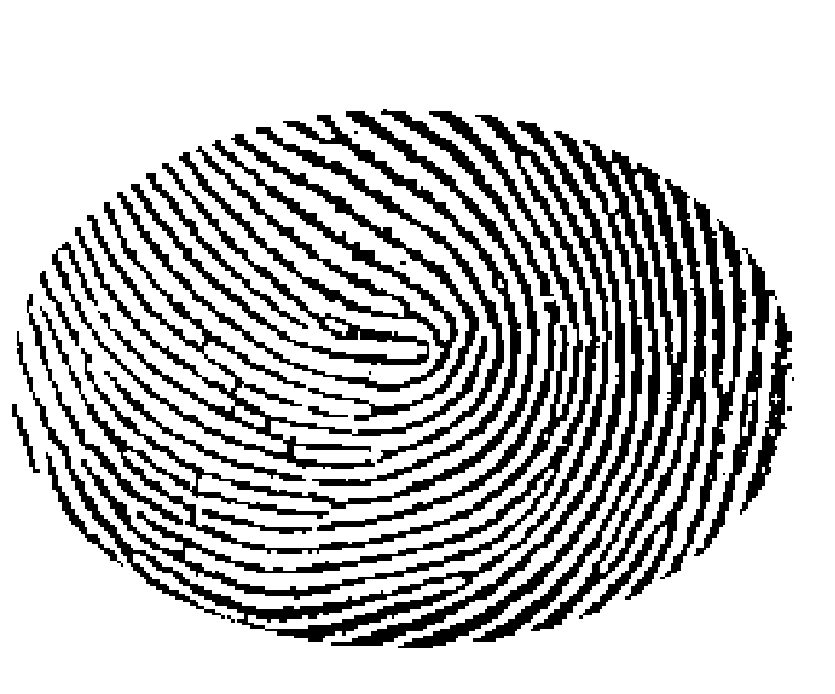}}
  \subfigure[Blurred image]{\includegraphics[scale=0.23]{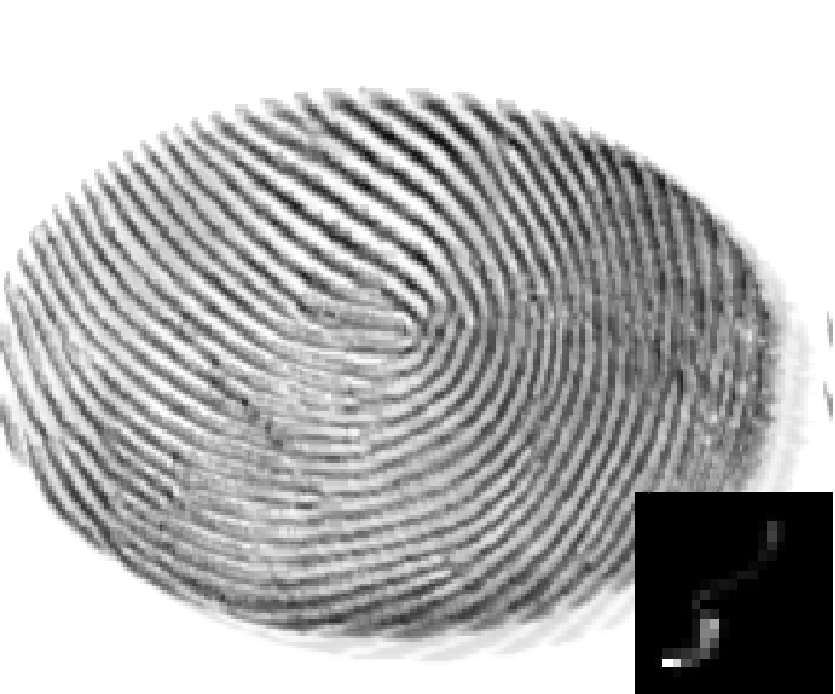}}
  \subfigure[Pan et al. \cite{pan2016l_0}]{\includegraphics[scale=0.23]{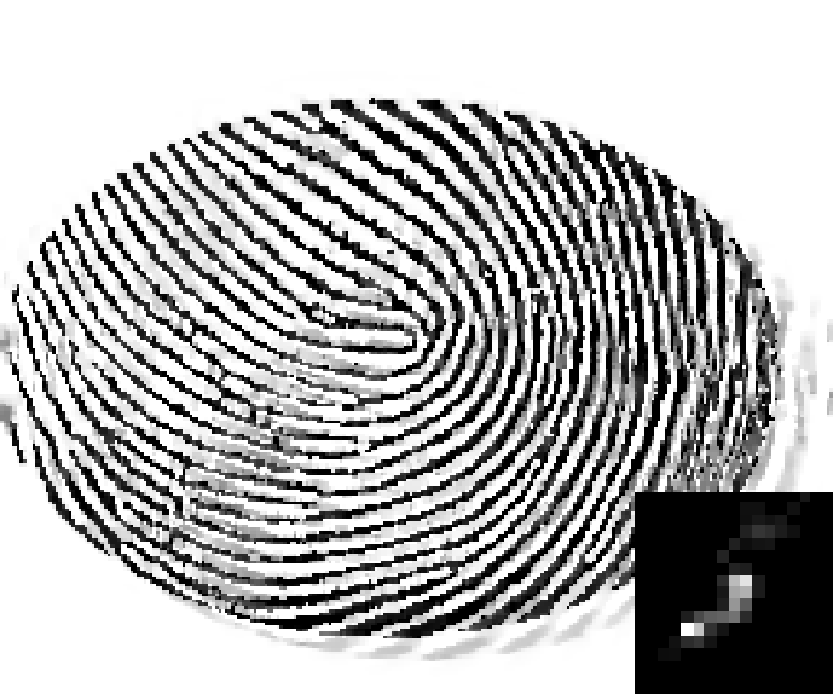}}
  \subfigure[Wen et al. \cite{wen2020simple}]{\includegraphics[scale=0.23]{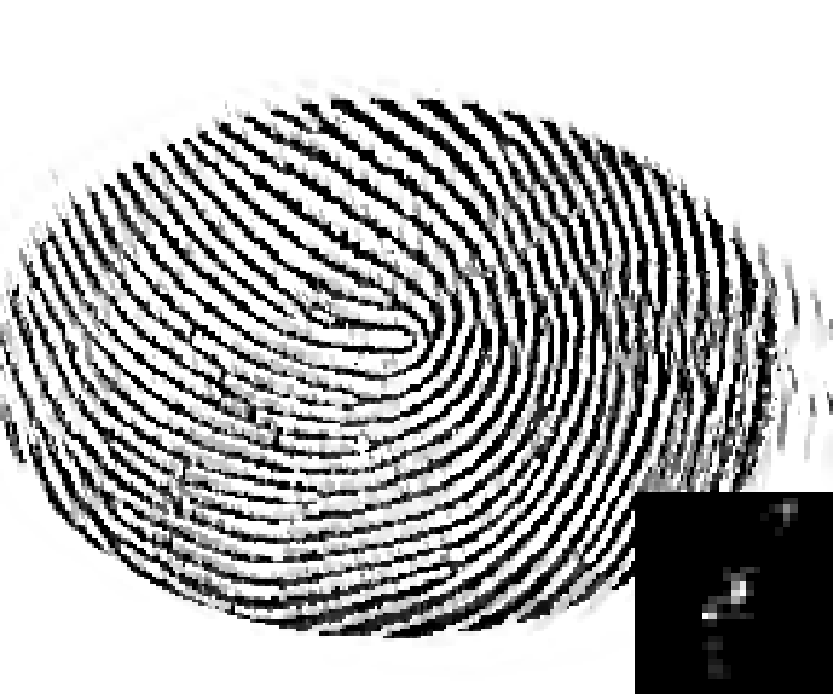}}
  \subfigure[Lv et al. \cite{lv2021blind}]{\includegraphics[scale=0.23]{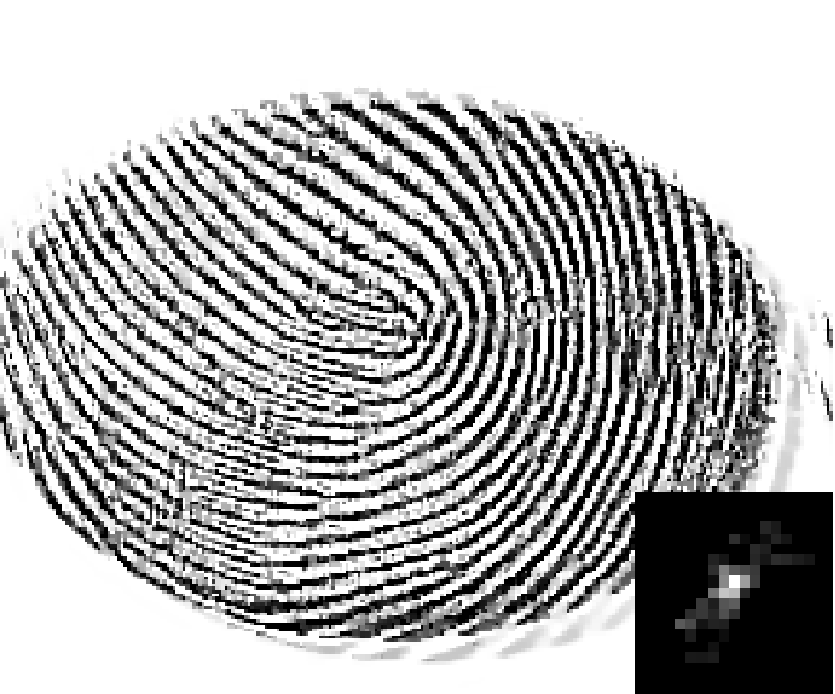}}
  \subfigure[Ours]{\includegraphics[scale=0.23]{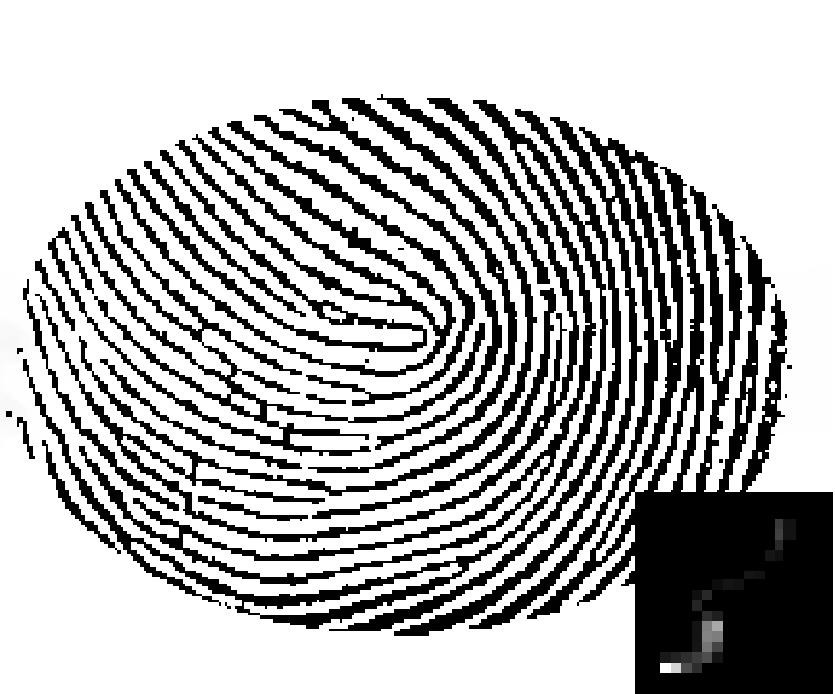}}
  \caption{A visual comparison of \textit{binary} fingerprint images. (c) Pan et al.'s method \cite{pan2016l_0} emphasizes the sparsity prior of intensity in the latent image, while (d) Wen et al.'s method \cite{wen2020simple} focuses on the extreme channel prior, and (e) Lv et al.'s method \cite{lv2021blind} leverages the binary intensity prior. It is evident that (f) our method outperforms these approaches by effectively reducing artifacts while better preserving structural details. (For improved visualization, the estimated kernels are displayed in the lower right corner of the final recovered images.) }
\end{figure}

%\newpage

It is noteworthy that in primary image deconvolution methods, image pixel values are generally treated as continuous variables  \cite{liu2021surface, lv2021blind}. 
However, in the field of digital image processing, images are acquired and stored in a digital format \cite{mei2015improving}. In particular, for several important classes of images, such as barcodes, text, and pattern images, their pixels are restricted to distinct value sets, as illustrated in Figure 1. For binary images (e.g., text and barcode images), pixel values are typically categorized into foreground and background categories. For pattern images, on the other hand, multiple distinct pixel values are used \cite{mei2015improving}.
In the context of imaging inverse problems, it is widely recognized that leveraging appropriate knowledge about the image can improve recovery performance. 
For example, numerical evidence (see Figure 2) shows that several existing advanced BID approaches \cite{wen2020simple, pan2016l_0} may produce unsatisfactory results when handling binary images with complex and repetitive structures, such as fingerprint images, because of inadequate blur kernel estimation.
As previously highlighted, specific image types, such as binary and pattern images, exhibit limited pixel intensity ranges. Therefore, it is reasonable to expect that accurately estimating and effectively utilizing these pixel values during the optimization process can provide valuable information for enhancing recovery performance.

%\newpage

Taking into account the distribution properties of image pixels, this paper investigates BID problems for specific binary and patterned images, introducing a novel regularizer that constrains distinct pixel intensities. In summary, the main contributions of this work are as follows.

\begin{itemize}
\item Building upon the distribution characteristics of image pixel intensities, we propose a unified, simple, yet effective model tailored for blind binary and pattern image deblurring. This model incorporates the pixel intensity constraint into the optimization process, resulting in significantly improved recovery performance.

\item Differing from mainstream model-driven BID methods, the proposed algorithm eliminates the necessity for additional non-blind deconvolution implementation, thereby streamlining the overall deblurring process.

\item Numerical experiments demonstrate the effectiveness of our proposed algorithm in restoring both binary images with two values and multi-valued pattern images. The recovery results show that our algorithm provides a noticeable boost compared to many relevant BID approaches in both quantitative and qualitative evaluations.
\end{itemize}

The remainder of this paper is organized as follows. Section 2 provides a brief review of related works. In Section 3, we introduce the pixel intensity constraint and present the proposed model as well as its optimization schemes. In Section 4, we compare the proposed algorithm with several related popular BID approaches, followed by a thorough analysis of the experimental results. The advantages and disadvantages of the proposed model and algorithm are analyzed in Section 5. Finally, the concluding remarks and future research directions are given in Section 6.

\section{Related work}

This section provides a concise overview of several prevailing BID methods, focusing primarily on the model-based and learning-based approaches.

\textbf{Model-based methods.} Fergus et al. \cite{fergus2006removing} exploited the heavy-tailed distribution of natural images' gradient histograms and the sparsity of blur kernels. However, it has been demonstrated that blind deconvolution methods based on image gradients tend to favor blurry images over sharp ones \cite{krishnan2011blind}. To address this issue, several other natural image priors have been introduced that prioritize sharp images over blurred ones.
Typically, Krishnan et al. \cite{krishnan2011blind} investigated the normalized sparsity prior for blind deconvolution. In \cite{xu2013unnatural}, Xu et al. developed an unnatural 
$\ell_0$ sparse expression for both uniform and non-uniform motion deblurring. Subsequently, Pan et al. \cite{pan2016blind} proposed an effective BID method based on the dark channel prior, which enforces the sparsity of the dark channels in the recovered images. Chen et al. \cite{chen2019blind} introduced a blind deblurring framework based on the local maximum gradient prior, where the maximum gradient value of a local patch diminishes during the blur process. Wen et al. \cite{wen2020simple} developed an efficient deblurring algorithm that incorporates the local minimal pixel prior, identifying local minima in pixel values within non-overlapping patches.
More recently, Liu et al. \cite{liu2021surface} explored the intrinsic geometric structure of intermediate images and proposed the surface-aware prior to effectively suppress undesired artifacts while simultaneously preserving sharp edges.

The assumption that pixels take on several distinct values has been explored in recent works for blind binary image deblurring. For instance, Pan et al. \cite{pan2016l_0} proposed an 
$\ell_0$ quasi-norm regularizer focusing on the zero peak in pixel values. The authors illustrated that incorporating additional intensity priors helps identify salient edges in text images, resulting in more accurate blur kernel estimation. However, a limitation of the 
$\ell_0$ intensity term is its fragility to binary images whose pixel values cluster around two peaks far from zero.
Subsequently, Liu et al. \cite{liu2018qr} introduced an alternating minimization approach for QR code image blind deblurring, where pixel values are constrained to two distinct classes: foreground and background pixels. To preserve the binary nature of barcode images, the authors incorporated a binarization regularizer into their minimization model. Similarly, Lv et al. \cite{lv2021blind} developed a sophisticated soft-rounding operator for blind binary image deblurring. During the image estimation stage, this soft-rounding function helps enforce the pixels in the estimated image to approach binary values.

Although recent literature has attempted to leverage pixel intensity properties to enhance BID performance for binary images, these approaches commonly encounter two main limitations. One limitation is that the developed binarization regularizers in these methods may not accurately characterize the properties of binary images, resulting in numerous pixel values in the recovered image deviating from the intended two distinct values and exhibiting a more complex distribution. Another limitation is that these binary BID methods cannot be straightforwardly applied or easily extended to address pattern images with multiple distinct pixel values.

\textbf{Learning-based methods.} Inspired by the unprecedented achievements of deep learning methods in the image processing community, a variety of learning-based approaches have also been proposed to address BID tasks. For example, Schuler et al. \cite{schuler2015learning} trained a deep network to estimate the blur kernel and then adopted a conventional non-blind deconvolution approach to recover the latent sharp image. In \cite{chakrabarti2016neural}, Chakrabarti used a neural network trained to predict the complex Fourier coefficients of a deconvolution filter, which can be applied to the input patch for restoration. Kupyn et al. \cite{kupyn2018deblurgan} introduced an end-to-end learning approach for deblurring using conditional generative adversarial networks (GANs). Ren et al. \cite{ren2020neural} proposed a neural blind deconvolution method called SelfDeblur, which leverages deep image priors and fully connected layers to effectively capture the prior information of both the clean images and the blur kernel.
%%Readers are referred to [32] for an extensive review of deep learning-based BID methods.

It is noteworthy that, despite achieving significant empirical success and producing highly competitive results, many data-driven deep learning methods suffer from several limitations. Firstly, the supervised nature of training deep networks demands a substantial dataset and depends heavily on the consistency between the training and test data for notable efficacy. Secondly, the lack of flexibility necessitates retraining the network to accommodate different degradation scenarios. Thirdly, these methods frequently require substantial computational resources and time, making them impractical in resource-constrained applications.
In this work, we mainly focus on model-based BID methods, aiming to preserve high flexibility while pursuing promising recovery performance.

\section{Our proposed model and algorithmic framework}
\subsection{Motivation}

As illustrated in Figure 1, binary and pattern images exhibit pixel intensities that belong to a very limited number of distinct values. This characteristic motivates us to leverage pixel intensity information during the iterative process to enhance recovery performance for such images. In the literature, Pan et al. \cite{pan2016l_0} introduced an 
$\ell_0$ quasi-norm intensity regularizer for blind text image deblurring, exploiting the sparsity prior of pixel intensities. However, this approach faces limitations when applied to binary images, and certainly for more complex multi-valued pattern images.
Given that binary images have only two possible values for each pixel, Lv et al. \cite{lv2021blind} proposed incorporating a rounding operator for blind binary image deblurring. However, this approach fails to  accurately  align the pixel intensity distributions between the estimated and ground truth images.

To address these limitations, our method proposes directly constraining the pixel intensities of the recovered image to a predetermined set consistent with the true pixel intensities of the ground truth. As a result, by integrating the pixel intensity constraint with the  $\ell_0$ quasi-norm gradient sparsity regularizer, the proposed unified framework not only precisely constrains the pixel intensities of binary and pattern images to specific subsets but also preserves the sharpness of the estimated images effectively.

\subsection{Pixel intensity constraint BID model}
Based on the distinctive pixel characteristics of binary and pattern images, we use the $\ell_0$ quasi-norm of image gradients and the $\ell_2$ norm of the blur kernel as regularization terms within a pixel intensity constrained MAP framework.
More specifically, we aim to address the following non-convex and constrained optimization problem for blind deconvolution of binary and pattern images.
\begin{equation}\label{eq: proposed constraint1 model}
\min _{\mathbf{x}_{i,j} \in \Omega,\mathbf{k}} \frac{1}{2}\|\mathbf{k*x}-\mathbf{y}\|_{F}^{2} + \lambda_{1}\|\nabla\mathbf{x}\|_{0} + \lambda_{2}\|\mathbf{k}\|_{F}^{2},
\end{equation}
where $\mathbf{x},\mathbf{y} \in \mathbb{R}^{m\times n}$ are the latent sharp image and observed blurry image, respectively. $\mathbf{k}$ is the unknown blur kernel. $\Omega=\{\alpha_1,\alpha_2,\ldots,\alpha_s\}$ denotes the set of estimated pixel intensities. 
It is noteworthy that in the context of blind binary image deconvolution \cite{lv2021blind}, the two pixel values of the ground truth image are assumed to be known in advance. In contrast, our proposed method for blind deconvolution of binary and pattern images focuses on approximating the pixel intensity set $\Omega$. This distinction also serves as a key highlight of our work. 
Additionally,  $\lambda_1$ and $\lambda_2$ are the two regularization parameters, 
the $\|\nabla\mathbf{x}\|_{0}$ is used to ensure the preservation of sharp edges, while $\|\mathbf{k}\|_{F}^{2}$ is utilized to impose smoothness and stability on the blur kernel.
The discrete gradient operator $\nabla: \mathbb{R}^{m\times n} \rightarrow (\mathbb{R}^{m\times n}, \mathbb{R}^{m\times n})$ is defined by
\[
(\nabla \mathbf{x})_{i,j} = \left((\nabla_1 \mathbf{x})_{i,j},(\nabla_2 \mathbf{x})_{i,j}\right)
\]
with
\[
(\nabla_1 \mathbf{x})_{i,j}=\left\{\begin{array}{ll}
\mathbf{x}_{i+1,j} - \mathbf{x}_{i,j}, & \mathrm{if} \   i < m, \\
\mathbf{x}_{1,j} - \mathbf{x}_{m,j}, & \mathrm{if} \  i = m,
\end{array}\right.
\]
and
\[
(\nabla_2 \mathbf{x})_{i,j}=\left\{\begin{array}{ll}
\mathbf{x}_{i,j+1} - \mathbf{x}_{i,j}, & \mathrm{if} \  j < n, \\
\mathbf{x}_{i,1} - \mathbf{x}_{i,n}, & \mathrm{if} \  i = n,
\end{array}\right.
\]
for $i=1,\ldots,m$ and $j=1,\ldots,n$, and $\mathbf{x}_{i,j}$ denotes the pixel intensity at position $(i,j)$ in the image.

For the sake of computational feasibility, we equivalently reformulate (\ref{eq: proposed constraint1 model}) as a non-constrained minimization problem.
\begin{equation}\label{eq: proposed non-constraint1 model}
\min _{\mathbf{k, x}} \frac{1}{2}\|\mathbf{k * x-y}\|_{F}^{2} +  \sum_{i, j=1}^{m, n} \mathcal{I}_{\Omega}\left(\mathbf{x}_{i, j}\right)+\lambda_{1}\|\nabla\mathbf{x}\|_{0}+\lambda_{2}\|\mathbf{k}\|_{F}^{2},
\end{equation}
where $\mathcal{I}_{\Omega}$ is the introduced \textit{indicator function} defined as:
\begin{equation}\label{eq: indicator function}
\mathcal{I}_{\Omega}(x)=\left\{\begin{array}{ll}
0, & x\in \Omega, \\
+ \infty, & x  \notin  \Omega.
\end{array}\right.
\end{equation}
To address the proposed model (\ref{eq: proposed non-constraint1 model}), we adopt a two-step iterative minimization approach that decouples image recovery from blur kernel estimation. Specifically, the solution to (\ref{eq: proposed non-constraint1 model}) is obtained by alternately solving the following two subproblems.
\begin{equation}\label{eq:image}
\min _{\mathbf{x}} \frac{1}{2}\|\mathbf{k * x-y}\|_{F}^{2} +  \sum_{i, j=1}^{m, n} \mathcal{I}_{\Omega} \left(\mathbf{x}_{i, j}\right)+\lambda_{1}\|\nabla\mathbf{x}\|_{0},
\end{equation}
and
\begin{equation}\label{eq:kernele}
\min _{\mathbf{k}}\frac{1}{2}\|\mathbf{x* k-y}\|_{F}^{2}+\lambda_{2}\|\mathbf{k}\|_{F}^{2}.
\end{equation}
In the intermediate image restoration stage, we employ the half-quadratic splitting (HQS) method, where each subproblem has a closed-form solution. In the blur kernel estimation stage, efficient computation can be achieved using FFTs.

\subsection{Latent image estimation}
To address the constrained non-convex optimization problem (\ref{eq:image}), we resort to the widely used HQS approach \cite{xu2011image} for its solution.
By introducing two auxiliary variables $\mathbf{u}=\mathbf{x},\mathbf{v}=\nabla\mathbf{x}$, the problem (\ref{eq:image}) can be reformulated as:
%%\begin{eqnarray}
%%\nonumber
%%\min _{\mathbf{x},\mathbf{u},\mathbf{v}} \frac{1}{2}\|\mathbf{k * x-y}\|_{F}^{2} +  \sum_{i, j=1}^{m, n} \mathcal{I}_{\Omega} \left(\mathbf{u}_{i, j}\right)+\lambda_{1}\|\mathbf{v}\|_{0}   &  & \\
%%+ \frac{\rho_1}{2}||\mathbf{u} - \mathbf{x}||_F^2 + \frac{\rho_2}{2}||\mathbf{v} - \nabla\mathbf{x}||_F^2,  & &
%%\end{eqnarray}
\begin{equation}
 \min_{\mathbf{x},\mathbf{u},\mathbf{v}} \frac{1}{2}\|\mathbf{k * x-y}\|_{F}^{2} +  \sum_{i, j=1}^{m, n} \mathcal{I}_{\Omega} \left(\mathbf{u}_{i, j}\right)+\lambda_{1}\|\mathbf{v}\|_{0}
+ \frac{\rho_1}{2}||\mathbf{u} - \mathbf{x}||_F^2 + \frac{\rho_2}{2}||\mathbf{v} - \nabla\mathbf{x}||_F^2,
\end{equation}
where $\rho_1$ and $\rho_2$ are the penalty parameters. Since the three variables $\mathbf{x},\mathbf{u},\mathbf{v}$ are coupled, we solve them alternatively.
\textbf{Algorithm 1}  summarizes the main steps for intermediate image estimation.
More concretely,

\textit{$\bullet$ $\mathbf{x}$ subproblem:}
\begin{equation}
 \min_{\mathbf{x}}\frac{1}{2}\|\mathbf{k * x-y}\|_{F}^{2}+\frac{\rho_{1}}{2}\|\mathbf{x}-\mathbf{u}\|_{F}^{2}+\frac{\rho_{2}}{2}\|\nabla\mathbf{x}-\mathbf{v}\|_{F}^{2}.
\end{equation}
According to the first-order optimization condition, also with the assumption of periodic boundary conditions on the blurry and difference, the closed-form solution can be computed by
\begin{equation}
\mathbf{x}=\mathcal{F}^{-1}\left(\frac{\overline{\mathcal{F}(\mathbf{k})} \circ \mathcal{F}(\mathbf{y})+\rho_{1} \mathcal{F}(\mathbf{u})+\rho_{2}\mathcal{F}_1}{\overline{\mathcal{F}\left(\mathbf{k}\right)} \circ \mathcal{F}(\mathbf{k})+\rho_{1} \mathcal{F}(\delta)+\rho_{2}\mathcal{F}_2}\right),
\end{equation}
where $\mathcal{F}_1 = \overline{\mathcal{F}\left(\nabla_{1}\right)} \circ \mathcal{F}\left(\mathbf{v}_{1}\right)+\overline{\mathcal{F}\left(\nabla_{2}\right)} \circ \mathcal{F}\left(\mathbf{v}_{2}\right)$ and $\mathcal{F}_2 =\overline{\mathcal{F}\left(\nabla_{1}\right)} \circ \mathcal{F}\left(\nabla_{1}\right)+\overline{\mathcal{F}\left(\nabla_{2}\right)} \circ \mathcal{F}\left(\nabla_{2}\right)$.  $\mathcal{F}(\cdot)$ and $\mathcal{F}^{-1}(\cdot)$ denotes the fast Fourier transforms (FFTs) and inverse FFTs, respectively. $\overline{\cdot}$ denote the conjugate operator, $\delta$ is the delta function, ``$\circ $'' and ``$- $'' denote the component-wise multiplication and division.

\textit{$\bullet$ $\mathbf{v}$ subproblem:}
\begin{equation}
\min_{\mathbf{v}}\frac{\rho_{2}}{2}\|\nabla\mathbf{x}-\mathbf{v}\|_{F}^{2}+\lambda_{1}\|\mathbf{v}\|_{0}.
\end{equation}
%We can efficiently obtain the closed-form solution of the optimization problem (12) by the following hard-thresholding operator.
The solution of (12) can be efficiently obtained by the following hard-thresholding operation.
\begin{equation}
\mathbf{v}_{i, j}=\left\{\begin{array}{ll}
(\nabla \mathbf{x})_{i, j}, & \left|(\nabla \mathbf{x})_{i, j}\right|^{2} \geq \frac{2\lambda_{1}}{\rho_{2}}, \\
0, & \text { otherwise. }
\end{array}\right.
\end{equation}

\textit{$\bullet$ $\mathbf{u}$ subproblem:}
\begin{equation}
\min_{\mathbf{u}}\frac{\rho_{1}}{2}\|\mathbf{x}-\mathbf{u}\|_{F}^{2} + \sum_{i, j=1}^{m, n} \mathcal{I}_{\Omega}\left(\mathbf{u}_{i, j}\right). \\
\end{equation}
%Both terms of equation (14) can be decomposed into the summation over each element of
%$\mathbf{u}$.
It can be seen that the above optimization problem can be addressed by solving for each element individually.
\begin{equation}
\min_{\mathbf{u}_{i,j}}\frac{\rho_{1}}{2}(\mathbf{x}_{i,j}-\mathbf{u}_{i,j})^{2}+\mathcal{I}_{\Omega}(\mathbf{u}_{i,j}).
\end{equation}
According to Theorem 3.1 below, the $\mathbf{u}$ subproblem can be obtained using the following formula.
\begin{equation}
\mathbf{u}_{i, j}=\left\{\begin{array}{ll}
\alpha_{1}, & \mathbf{x}_{i, j}<\frac{\alpha_1+\alpha_2}{2}, \\
\alpha_{k}, & \frac{\alpha_{k-1}+\alpha_{k}}{2} \leq \mathbf{x}_{i, j} < \frac{\alpha_{k}+\alpha_{k+1}}{2}, \  k \in \{2,\ldots,s-1\}\\
\alpha_{s}, & \mathbf{x}_{i, j}\geq \frac{\alpha_{s-1}+\alpha_{s}}{2}.
\end{array}\right.
\end{equation}\par
\noindent
\textbf{Theorem 3.1} Let $y \in \mathbb{R}$ and consider a minimization problem of the form
\begin{equation}
x^{\ast}= \mathrm{arg} \min_{x}  \frac{\mu}{2}(x-y)^{2}     +    \mathcal{I}_{\Omega}(x),
\end{equation}
where $\mathcal{I}_{\Omega}$ is the indicator function defined as (\ref{eq: indicator function}). It can be derived that the optimal solution takes the following form.
\begin{equation}
x^{\ast}=\left\{\begin{array}{ll}
\alpha_{1}, & y<\frac{\alpha_1+\alpha_2}{2}, \\
\alpha_{k}, & \frac{\alpha_{k-1}+\alpha_{k}}{2} \leq y < \frac{\alpha_{k}+\alpha_{k+1}}{2}, \  k \in \{2,\ldots,s-1\}\\
\alpha_{s}, & y \geq \frac{\alpha_{s-1}+\alpha_{s}}{2}.
\end{array}\right.
\end{equation}\par
\noindent
\textbf{Proof:} 
%%Please refer to the Appendix A. $\blacksquare$
Without the loss of generality, we assume $\alpha_{1}<\alpha_{2}<\ldots<\alpha_{s}$.

\textit{Case 1:} When $\Omega=\{\alpha_{1},\alpha_{2}\}$, it is easy to derive that if $y<\frac{\alpha_1+\alpha_2}{2}$, it computes $(\alpha_1 - y)^2 < (\alpha_2 - y)^2$,
thus $x^{\ast}=\alpha_{1}$; if $y>\frac{\alpha_1+\alpha_2}{2}$, it obtains $(\alpha_1 - y)^2 > (\alpha_2 - y)^2$, thus $x^{\ast}=\alpha_{2}$; if $y=\frac{\alpha_1+\alpha_2}{2}$, it computes $(\alpha_1 - y)^2 = (\alpha_2 - y)^2$, thus $x^{\ast}=\alpha_{1}$ or $x^{\ast}=\alpha_{2}$.

\textit{Case 2:} When $\Omega=\{\alpha_{1},\alpha_{2}, \alpha_{3}\}$, it derives that if $y<\frac{\alpha_1+\alpha_2}{2}$, it computes $(\alpha_1 - y)^2 < (\alpha_2 - y)^2<(\alpha_3 - y)^2$, thus $x^{\ast}=\alpha_{1}$;
if $y =\frac{\alpha_{1}+\alpha_{2}}{2}$, it computes $(\alpha_1 - y)^2 = (\alpha_2 - y)^2<(\alpha_3 - y)^2$, thus $x^{\ast}=\alpha_{1}$ or $x^{\ast}=\alpha_2$;
if $\frac{\alpha_{1}+\alpha_{2}}{2} < y<\frac{\alpha_2+\alpha_3}{2}$,
it computes $(\alpha_2 - y)^2 < (\alpha_1 - y)^2<(\alpha_3 - y)^2$, thus $x^{\ast}=\alpha_{2}$;
if $y = \frac{\alpha_2+\alpha_3}{2}$, it computes $(\alpha_2 - y)^2 = (\alpha_3 - y)^2<(\alpha_1 - y)^2$, thus $x^{\ast}=\alpha_{2}$ or $x^{\ast}=\alpha_3$;
if $y > \frac{\alpha_2+\alpha_3}{2}$, it computes $(\alpha_3 - y)^2 < (\alpha_2 - y)^2<(\alpha_1 - y)^2$, thus $x^{\ast}=\alpha_3$.

\textit{Case 3:} For the general cases $\Omega=\{\alpha_{1},\alpha_{2},\cdot\cdot\cdot,\alpha_{s}\}$, when $y$ falls within the interval $[\alpha_{k-1}, \alpha_{k}] \cup [\alpha_{k}, \alpha_{k+1}]$, $k \in \{2,\ldots,s-1\}$, obviously,  the optimal value can only be selected from the three values $\{\alpha_{k-1},\alpha_{k},\alpha_{k+1}\}$ and it will be reduced to the above-mentioned \textit{case 2}. Furthermore, in the situations where $y<\alpha_1$ ($y>\alpha_s$), it is easy to obtain $x^{\ast}=\alpha_{1}$ ($x^{\ast}=\alpha_{s}$), respectively. This completes the proof. $\blacksquare$

\subsection{Blur kernel estimation}
According to \cite{xu2013unnatural, cho2009fast, levin2011efficient}, the accuracy of blur kernel estimation based on intensity values is often inadequate. To improve the stability and precision of this process, one can achieve better results by utilizing image gradients rather than intensities to solve for $\mathbf{k}$.
As a result, we replace (\ref{eq:kernele}) with the following gradient-based formulation.
\begin{equation}\label{eq:ker}
\min _{\mathbf{k}} \frac{1}{2}\|\nabla \mathbf{x * k}-\nabla \mathbf{y}\|_{F}^{2}+\lambda_{2}\|\mathbf{k}\|_{F}^{2} ,
\end{equation}
Then, the solution of (\ref{eq:ker}) can be obtained through the FFTs.
\begin{equation}
\mathbf{k}=\mathcal{F}^{-1}\left(\frac{\overline{\mathcal{F}\left(\nabla_{1} \mathbf{x}\right)} \circ \mathcal{F}\left(\nabla_{1} \mathbf{y}\right)+\overline{\mathcal{F}\left(\nabla_{2} \mathbf{x}\right)} \circ \mathcal{F}\left(\nabla_{2} \mathbf{y}\right)}{\overline{\mathcal{F}\left(\nabla_{1} \mathbf{x}\right)} \circ \mathcal{F}\left(\nabla_{1} \mathbf{x}\right)+\overline{\mathcal{F}\left(\nabla_{2} \mathbf{x}\right)} \circ \mathcal{F}\left(\nabla_{2} \mathbf{x}\right)+\lambda_{2} \mathcal{F}(\delta)}\right) .
\end{equation}
After getting the blur kernel $\mathbf{k}$, we adjust it by zeroing any negative elements and normalizing to ensure that the sum equals $1$.
\textbf{Algorithm 2} describes the main steps for blur kernel estimation.  \par
\noindent
\textbf{Remark:} Following \cite{lv2021blind,pan2016l_0}, and considering the highly non-convex nature of problem (\ref{eq: proposed non-constraint1 model}), we also adopt a coarse-to-fine scheme to approximate the latent images and blur kernels.
Initially, an image pyramid $\mathbf{x}_{1}, \mathbf{x}_{2}, \ldots ,\mathbf{x}_{L}$ is constructed from the blurred image $\mathbf{x}$, where $\mathbf{x_1}=\mathbf{x}$ and $\mathbf{x}_{L}$ represents the most down-sampled version of $\mathbf{x}$.
The iterative process,  involving both the kernel and intermediate latent image, is initiated from the coarsest level $L$. Subsequently, the estimated blur kernel $\mathbf{k}_{L}$ is up-sampled and passed as initialization to level $L-1$.
Finally, through this progressive refinement, the blur kernel $\mathbf{k}$ and latent image $\mathbf{x}$ will be obtained at its finest level (i.e.,  the level 1).

\begin{algorithm}[!h]
    \caption{Intermediate image estimation stage.}
    \label{alg:AOS}
    \renewcommand{\algorithmicrequire}{\textbf{Input:}}
    \renewcommand{\algorithmicensure}{\textbf{Output:}}
    \begin{algorithmic}[1]
        \REQUIRE $\mathbf{y}$,  $\mathbf{k}$;  $\lambda_{1}, \rho_{1}=16\lambda_{1}$, $max_{\rho_{1}}$ and $max_{\rho_{2}}$. %%input
        \ENSURE Latent restoration image $\mathbf{x}$.    %%output
        \WHILE{$\rho_{1}\leq max_{\rho_{1}}$}
            \STATE updating $\mathbf{u}$ via (16).
            \STATE $\rho_{2}=2\lambda_{1}$;
            \WHILE{$\rho_{2}\leq max_{\rho_{2}}$}
            \STATE updating $\mathbf{v}$ via (13).
            \STATE updating $\mathbf{x}$ via (11).
            \STATE $\rho_{2}=2\rho_{2}$;
            \ENDWHILE
            \STATE $\rho_{1}=2\rho_{1}$;
        \ENDWHILE
    \end{algorithmic}
\end{algorithm}

%\begin{algorithm}[!h]
%    \caption{Blur kernel estimation stage.}
%    \label{alg:AOS}
%    \renewcommand{\algorithmicrequire}{\textbf{Input:}}
%    \renewcommand{\algorithmicensure}{\textbf{Output:}}
%    \begin{algorithmic}[1]
%        \REQUIRE  $\mathbf{y}$;  $\lambda_{2}$, $\iota$ and $\kappa$. %%input
%        \ENSURE Blur kernel $\mathbf{k}$.    %%output
%        \WHILE{the stopping criterion is not satisfied}
%            \STATE solving $\mathbf{x}$ via Algorithm 1.
%            \STATE solving $\mathbf{k}$ via (20).
%            \STATE $\lambda_{2}=\max\{\lambda_{2}/\iota,\kappa\}$.
%        \ENDWHILE
%    \end{algorithmic}
%\end{algorithm}

\begin{algorithm}[!h]
    \caption{Blur kernel estimation stage.}
    \label{alg:AOS}
    \renewcommand{\algorithmicrequire}{\textbf{Input:}}
    \renewcommand{\algorithmicensure}{\textbf{Output:}}
    \begin{algorithmic}[1]
        \REQUIRE  $\mathbf{y}$;  $\lambda_{2}$, $\iota$, $\kappa$ and $N$. %%input
        \ENSURE Blur kernel $\mathbf{k}$.    %%output
        \FOR{$i=1,2\ldots,N$}
            \STATE solving $\mathbf{x}$ via Algorithm 1.
            \STATE solving $\mathbf{k}$ via (20).
            \STATE $\lambda_{2}=\max\{\lambda_{2}/\iota,\kappa\}$.
        \ENDFOR
    \end{algorithmic}
\end{algorithm}

\begin{figure}
  \centering
  %\subfigure[Intensity histogram]{\includegraphics[scale=0.42]{spixel.eps}}
  \subfigure[Binary image $\mathbf{x}^{\mathrm{ref}}$ by \cite{ren2020neural}]{\includegraphics[scale=0.35]{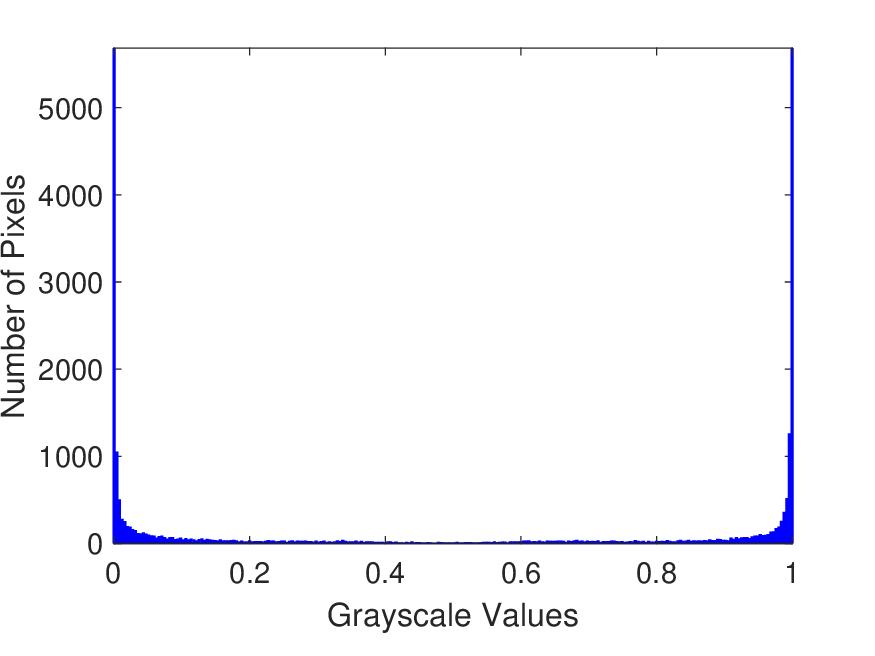}}
  \subfigure[Pattern image $\mathbf{x}^{\mathrm{ref}}$ by  \cite{pan2016l_0}]{\includegraphics[scale=0.35]{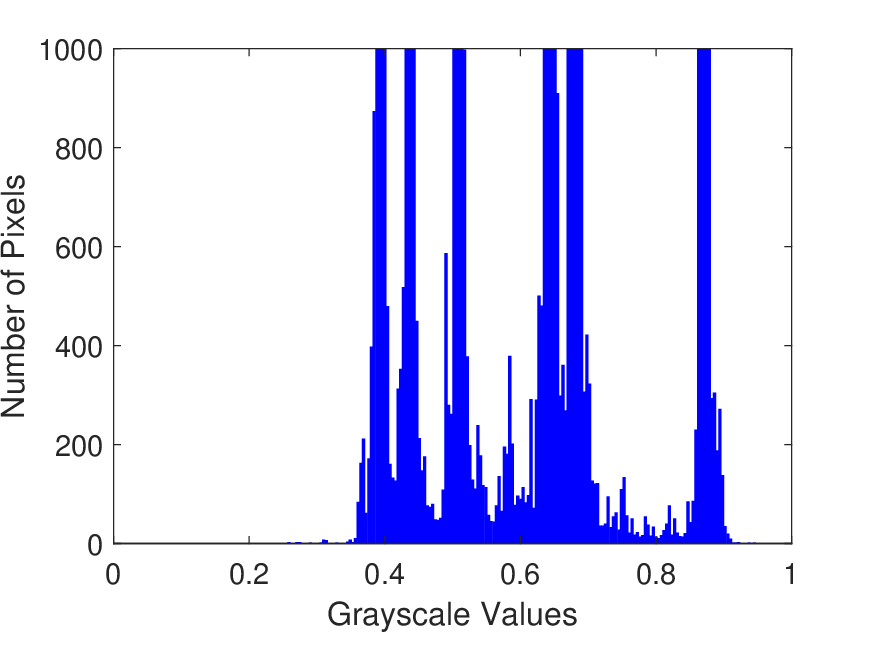}}
  \caption{An illustrative example of the pixel intensity distribution of $\mathbf{x}^{\mathrm{ref}}$. The pixel intensity histograms of (a) binary reference image $\mathbf{x}^{\mathrm{ref}}$ (recovered by \cite{ren2020neural}) and (b) pattern reference image $\mathbf{x}^{\mathrm{ref}}$  (recovered by \cite{pan2016l_0}).
%  The clustering number  $s$ (i.e., the number of distinct pixel values ââin the image)
The clustering number $s$ (i.e., the number of different pixel values in the image) can be approximated by the number of peaks in the histogram. }
\end{figure}

%\begin{figure*}[htbp]
%\centering
%\subfigure[Binary $\mathbf{x}^{\mathrm{ref}}$ obtained from \cite{ren2020neural}]{\includegraphics[width=0.40\textwidth]{rbinary.eps}}
%\subfigure[Pattern $\mathbf{x}^{\mathrm{ref}}$ obtained from  \cite{pan2016l_0}]{\includegraphics[width=0.40\textwidth]{rpattern.eps}}
%\caption{ An illustrative example of the pixel intensity property of $\mathbf{x}^{\mathrm{ref}}$. The pixel intensity histograms of (a) binary reference image $\mathbf{x}^{\mathrm{ref}}$ (recovered by \cite{ren2020neural}) and (b) pattern reference image $\mathbf{x}^{\mathrm{ref}}$  (recovered by \cite{pan2016l_0}).
%The clustering number  $s$ (i.e, the number of distinct pixel values ââ in the image) can be approximated by manually counting the number of peaks in the histogram. }
%\end{figure*}

\subsection{Distinct pixel values estimation}
There remains only one critical issue unresolved, that is,
how to determine the set $\Omega$ of distinctive pixel intensities $\{\alpha_{1},\alpha_{2},\cdot\cdot\cdot,\alpha_{s}\}$ in (4).
The accuracy of detected $\Omega$ has an important impact on our recovery performance.
However,  this is a challenging problem and obtaining $\Omega$  directly from the input blurry images $\mathbf{y}$ will lead to useless information.
As an alternative, we propose a simple yet effective strategy to obtain the $\Omega$ in this preliminary work.
Initially, we resort to an off-the-shelf BID approach to restore a relatively high-quality image (e.g.,  guided by experiments, using Ren et al.  \cite{ren2020neural} to restore binary images and Pan et al.  \cite{pan2016l_0} to restore pattern images, respectively), and the output is acted as a reference image $\mathbf{x}^{\mathrm{ref}}$.
Afterwards, we employ the K-Means clustering technique, to extract the sorted pixel intensities on this reference image $\mathbf{x}^{\mathrm{ref}}$.
The number of clusters $s$ can be manually determined by examining the intensity histogram of the reference image, see an illustrative example shown in Figure 3.
Obviously, as the number of different pixel values in the underlying image (i.e., number of clusters $s$)  increases, our strategy may become more susceptible to errors.
It is noted that although potential errors in the estimated pixel values may inevitable, our proposed algorithm still significantly enhance the recovery performance, as demonstrated by extensive empirical findings.

\section{Numerical experiments}

\subsection{Experiments setting}
We demonstrate the effectiveness of the proposed method through numerical experiments on three synthetic datasets.
The first dataset consists of 64 blurred binary images, generated from 8 clean binary images depicted in Figure 4(a-h) and 8 blur kernels as shown in Figure 5, based on the research conducted by Levin et al. \cite{levin2011understanding}.
The second dataset contains 32 blurred pattern images, generated from 4 multi-valued pattern images displayed in Figure 4(i-l) and the same 8 blur kernels from Levin et al. \cite{levin2011understanding}.
For the third dataset, we first apply the blur to the original images in Figure 4(a-d) using the kernels shown in Figure 5(a-d), respectively. Next, noise with a Blurred  Signal-to-Noise Ratio (BSNR) level of 40 dB is added to the blurred images. 
Similarly, the blur is applied again to the original images in Figure 4(e-h) using the kernels shown in Figure 5(e-h), respectively, and noise with a BSNR level of 30 dB is added to these blurred images.
This process generates the third dataset, consisting of eight blurry and noisy binary images.
The Peak Signal-to-Noise Ratio (PSNR) and the Structural Similarity Index Metric (SSIM) \cite{wang2004image} are employed as two objective metrics to evaluate the performance of competing algorithms. In general, higher PSNR and SSIM values indicate better recovery performance.

\begin{figure}
  \centering
  \subfigure[im1]{\includegraphics[scale=0.3]{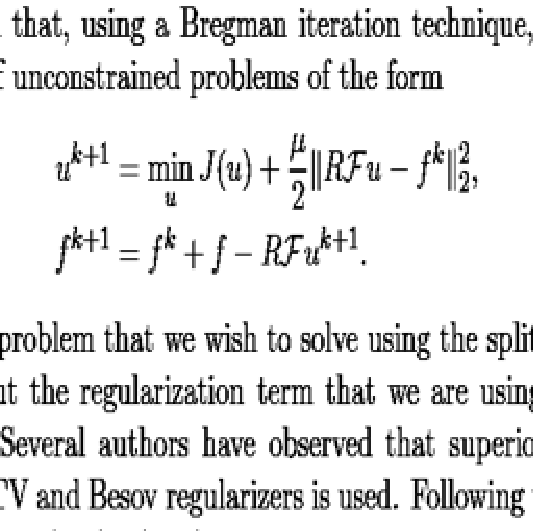}}
  \subfigure[im2]{\includegraphics[scale=0.3]{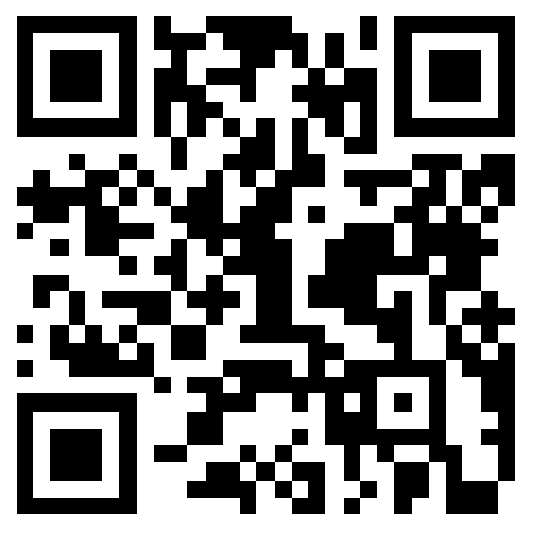}}
  \subfigure[im3]{\includegraphics[scale=0.3]{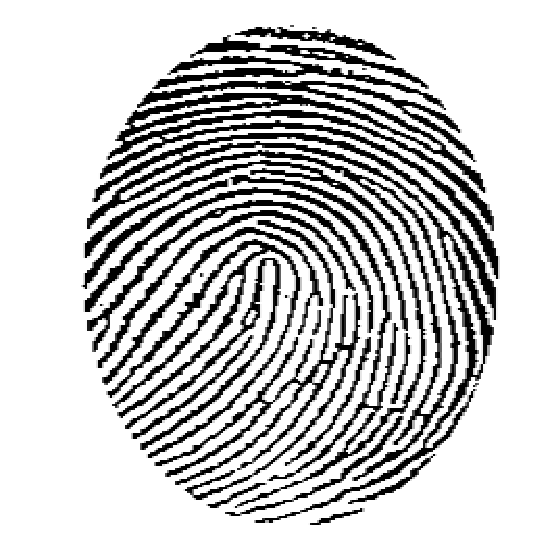}}
  \subfigure[im4]{\includegraphics[scale=0.3]{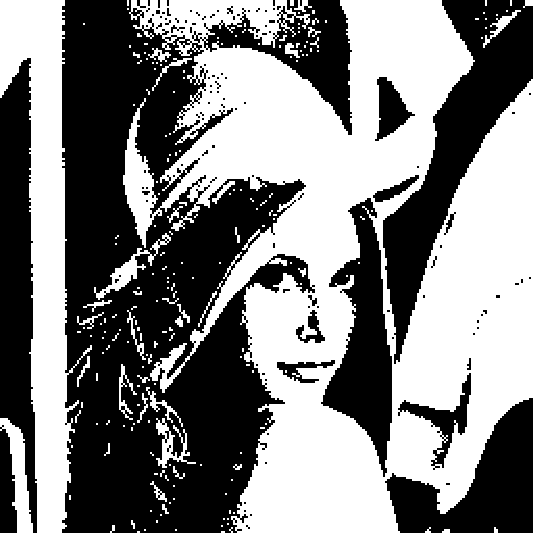}}
  \subfigure[im5]{\includegraphics[scale=0.3]{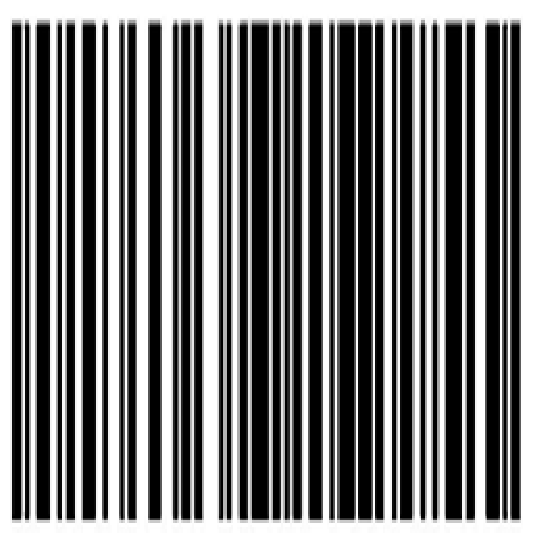}}
  \subfigure[im6]{\includegraphics[scale=0.3]{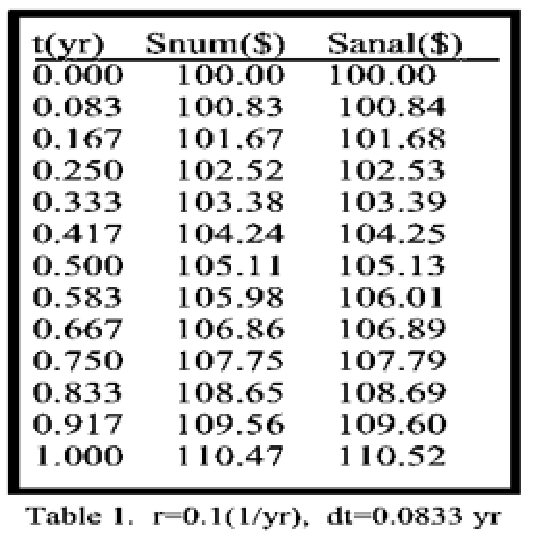}}
  \subfigure[im7]{\includegraphics[scale=0.3]{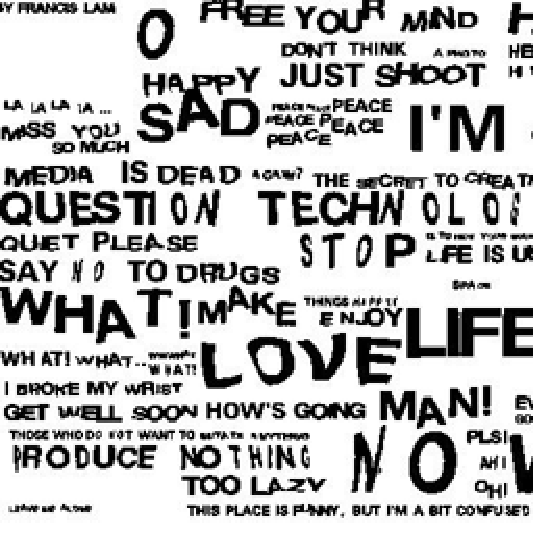}}
  \subfigure[im8]{\includegraphics[scale=0.3]{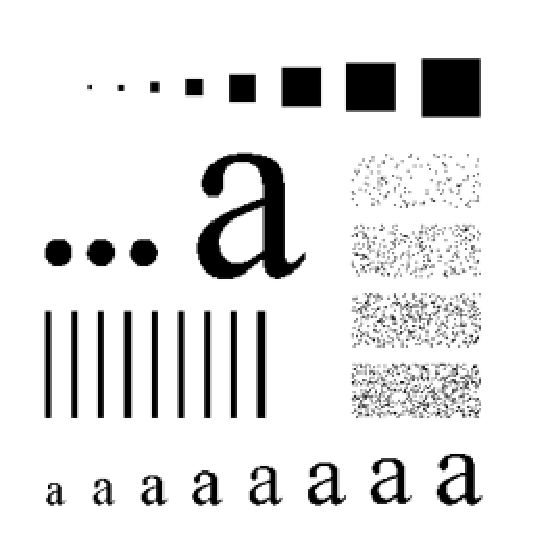}}
  \subfigure[im9]{\includegraphics[scale=0.3]{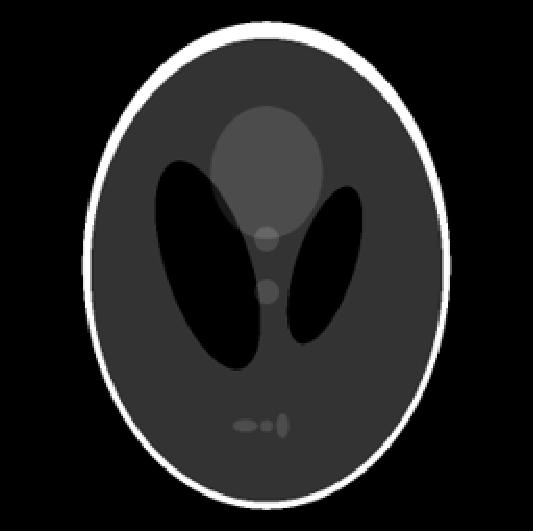}}
  \subfigure[im10]{\includegraphics[scale=0.3]{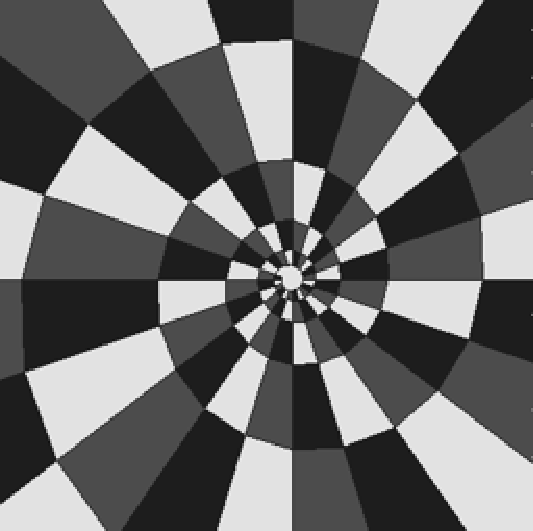}}
  \subfigure[im11]{\includegraphics[scale=0.3]{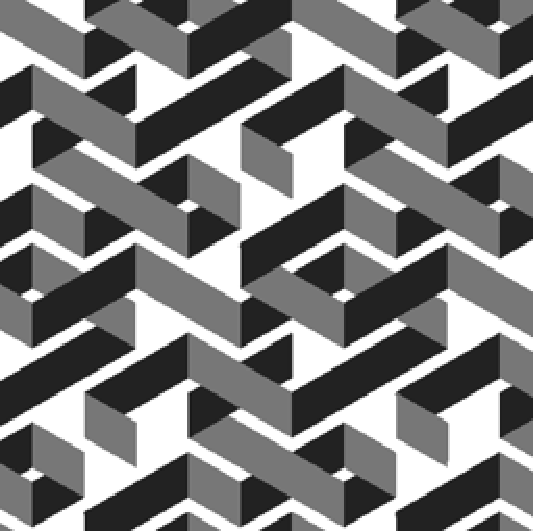}}
  \subfigure[im12]{\includegraphics[scale=0.3]{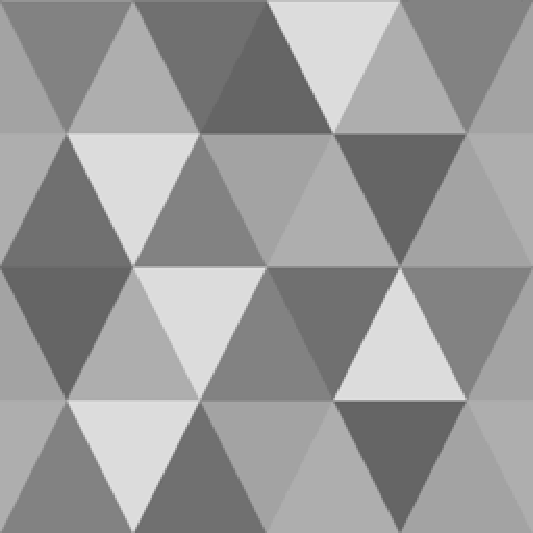}}
  \caption{All test images: eight binary images (a-h) and four pattern images (i-l).}%(a) im1, (b) im2, (c) im3, (d) im4, (e) im5, (f) im6, (g) im7, (h) im8.
\end{figure}

\begin{figure}
  \centering
  \subfigure[ker1]{\includegraphics[scale=0.5]{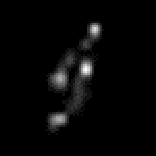}}
  \subfigure[ker2]{\includegraphics[scale=0.5]{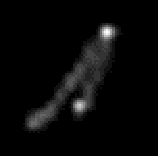}}
  \subfigure[ker3]{\includegraphics[scale=0.5]{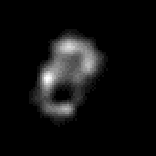}}
  \subfigure[ker4]{\includegraphics[scale=0.5]{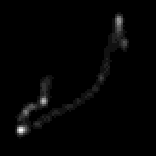}}
  \subfigure[ker5]{\includegraphics[scale=0.5]{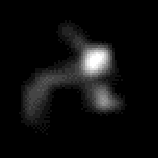}}
  \subfigure[ker6]{\includegraphics[scale=0.5]{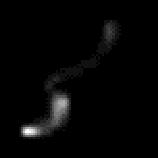}}
  \subfigure[ker7]{\includegraphics[scale=0.5]{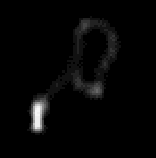}}
  \subfigure[ker8]{\includegraphics[scale=0.5]{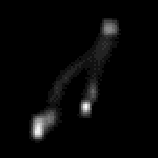}}
  \caption{All test kernels are from the research conducted by Levin et al. \cite{levin2011understanding}.}%(a) im1, (b) im2, (c) im3, (d) im4, (e) im5, (f) im6, (g) im7, (h) im8.
\end{figure}

%\newpage
\subsection{Implementation details}

Throughout the experiments, we set $\lambda_{1}=2e^{-3}, max_{\rho_{1}}=8, max_{\rho_{2}}=2e^{5}$ for the intermediate image estimation in Algorithm 1.
For blur kernel estimation, we set $\iota=1.1, \kappa=2e^{-4}, \lambda_{2}=2$, and the maximum number of iterations allowed,  $N=5$, in Algorithm 2.
For all the competing methods, parameters were either manually tuned to attain optimal performance or set according to recommended settings in the reference papers and codes.
Moreover, we validate the effectiveness of the proposed algorithm in two parts: 
(1) Following the experimental setting in \cite{lv2021blind,mei2015improving}, we assume the distinct true pixel values are ideally known; 
(2) Since exact pixel values are commonly unattainable in practice, we also provide the recovered results using predicted pixel intensities.
Correspondingly, we use the term ``ours'' to refer to the incorporation of estimated
$\Omega$  throughout the computation process, while ``ours*'' denotes the utilization of exact  $\Omega$ during the computation process. Similarly, ``Lv'' and ``Lv*'' indicate the use of estimated and exact $\Omega$ in their respective models.
All experiments were conducted in Matlab R2021b on a PC computer equipped with an AMD Ryzen 7 5800H 8-Core Processor (3.20 GHz) and 16 GB of memory.

\subsection{Blind binary image deconvolution}
On the first synthetic dataset consisting of 64 blurry binary images, 
we compare with several competing methods \cite{pan2016l_0,chen2019blind,ren2020neural,
wen2020simple,lv2021blind}.  As shown in Table 1 and Figure 6, our method performs favorably against all competing methods.
Particularly, it has an impressive average PSNR/SSIM values improvement up to (18.1 dB/0.0631), (22.27 dB/0.079), (21.94 dB/0.0882), (21.14 dB/0.0761), (22.38 dB/0.0772)  over Ren \textit{et al.} \cite{ren2020neural}, Chen et al.  \cite{chen2019blind}, Wen et al. \cite{wen2020simple}, Lv* et al.  \cite{lv2021blind}, Pan  et al.  \cite{pan2016l_0}, respectively.
Figure 7 and Figure 8 display the recovered results obtained by different methods for the QR code image ``im2'' and the number table image ``im6''.
From the enlarged window, it can be observed that the images restored by competing algorithms contain noticeable out-of-range pixel values, resulting in unpleasant artifacts in the recovered image.
In contrast, the proposed method significantly eliminates these unappealing artifacts and well preserves sharp edges by adding a constraint on the distribution of pixel intensities, ultimately yielding visually much superior results.

\begin{figure}
  \centering
  \subfigure[Average PSNR values]{\includegraphics[scale=0.38]{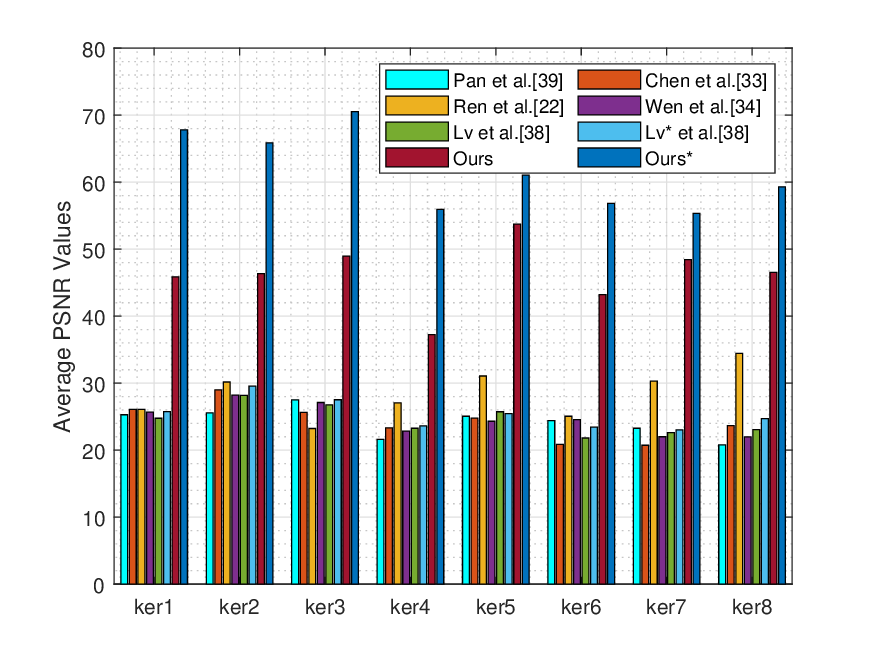}}
  \subfigure[Average SSIM values]{\includegraphics[scale=0.38]{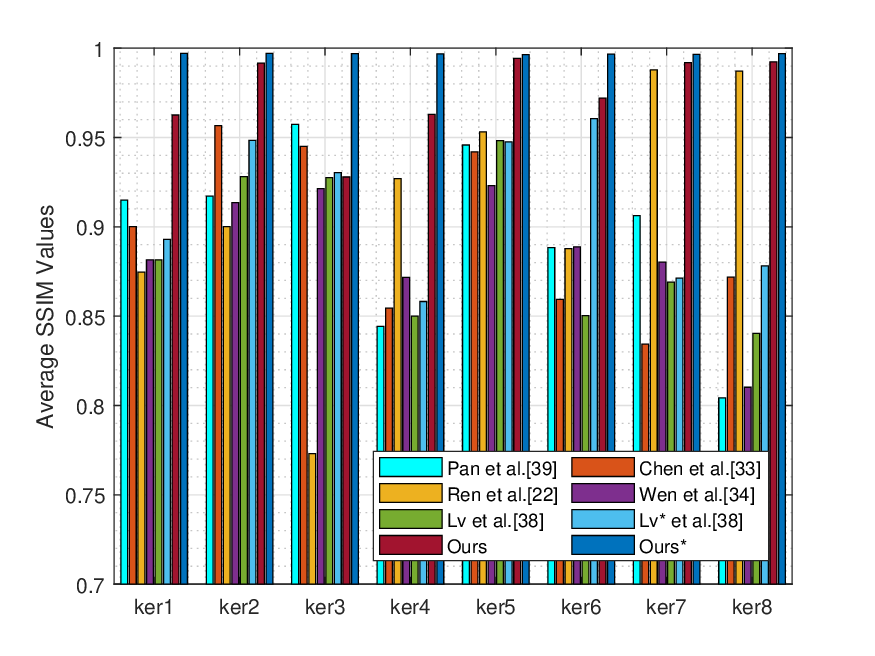}}
  \caption{Quantitative evaluations on 8 binary images of different blur kernels, it can be observed that our method performs competitively against other competing methods. }
\end{figure}
\begin{table}
\centering
\caption{Quantitative results (average PSNR/SSIM) of binary image deblurring.}
\footnotesize
\setlength{\tabcolsep}{11mm}{
\begin{tabular}{ccc}
\toprule
Method&PSNR $\uparrow$ &SSIM  $\uparrow$  \\
\midrule
Pan et al. \cite{pan2016l_0}&24.15&0.8973\\
Chen et al. \cite{chen2019blind}&24.26&0.8955\\
Ren et al. \cite{ren2020neural}&28.43&0.9114\\
Wen et al. \cite{wen2020simple}&24.59&0.8863\\
Lv et al. \cite{lv2021blind}&24.52&0.8869\\
Lv* et al. \cite{lv2021blind}&25.39&0.8984\\
Ours&46.53&0.9745\\
Ours*&61.58&0.9968\\
\bottomrule
\end{tabular}}
\end{table}
\begin{figure}
  \centering
  \subfigure[\scriptsize{Ground truth}]{\includegraphics[scale=0.33]{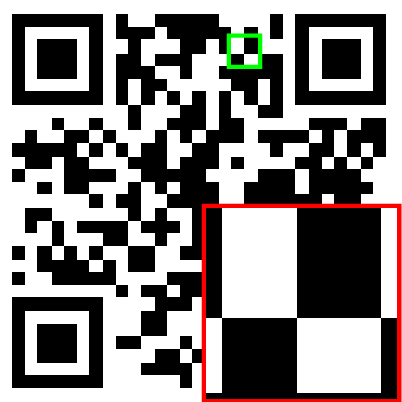}}
  \subfigure[\scriptsize{Blurry image}]{\includegraphics[scale=0.33]{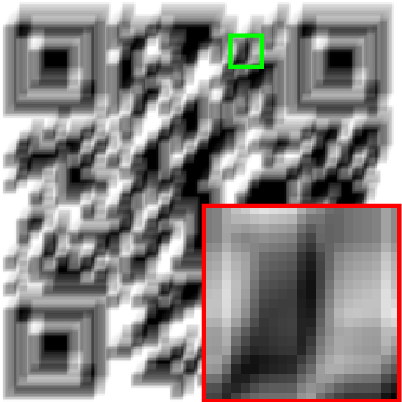}}
  \subfigure[\scriptsize{19.06/0.6934}]{\includegraphics[scale=0.33]{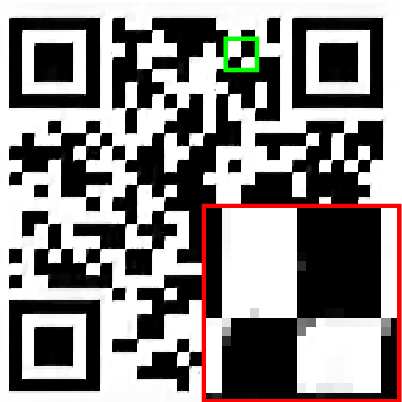}}
  \subfigure[\scriptsize{26.84/0.9072}]{\includegraphics[scale=0.33]{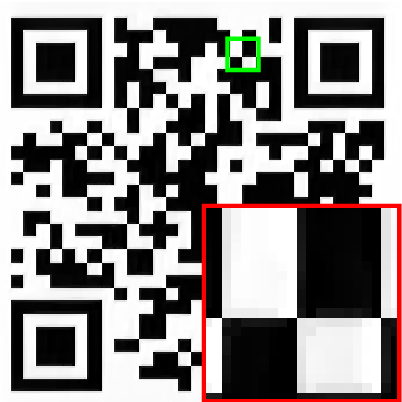}}
  \subfigure[\scriptsize{58.86/0.9999}]{\includegraphics[scale=0.33]{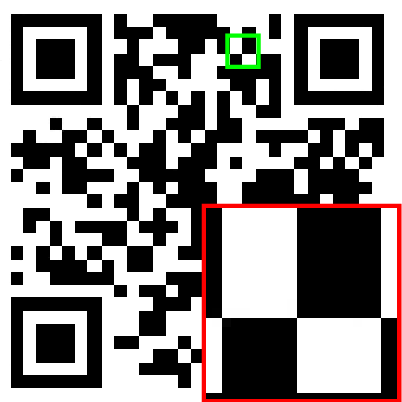}}
  \subfigure[\scriptsize{28.26/0.9152}]{\includegraphics[scale=0.33]{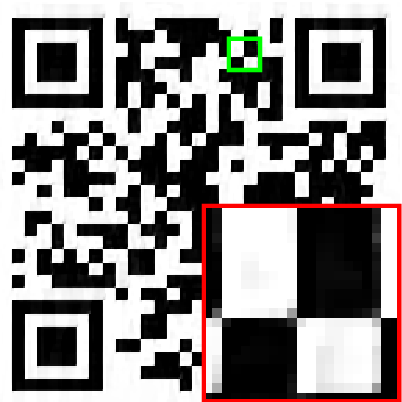}}
  \subfigure[\scriptsize{22.58/0.8252}]{\includegraphics[scale=0.33]{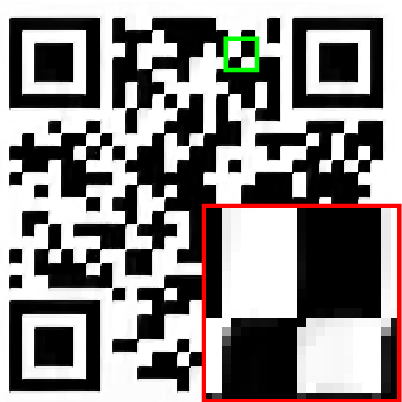}}
  \subfigure[\scriptsize{27.88/0.9367}]{\includegraphics[scale=0.33]{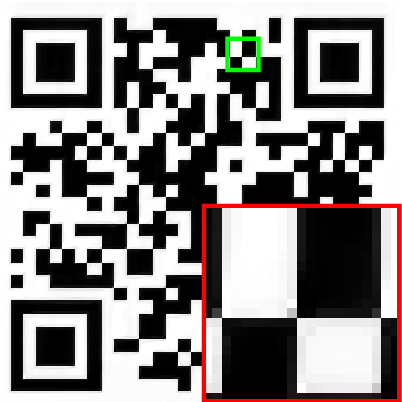}}
  \subfigure[\scriptsize{68.02/0.9999}]{\includegraphics[scale=0.33]{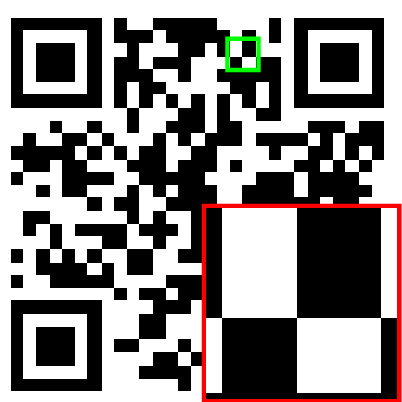}}
  \subfigure[\scriptsize{68.81/0.9999}]{\includegraphics[scale=0.33]{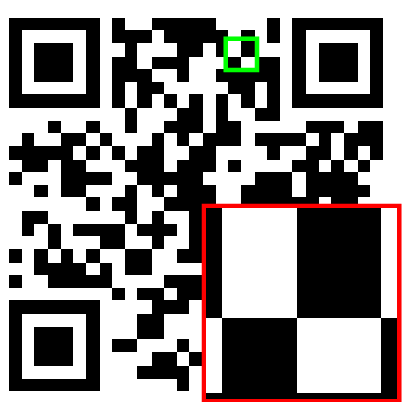}}
  \caption{The visual quality and quantitative results (PSNR/SSIM)  of different methods on ``im2''. (a) Ground truth, (b) Blurred image, (c) Pan  et al.  \cite{pan2016l_0},  (d) Chen et al. \cite{chen2019blind}, (e) Ren et al. \cite{ren2020neural}, (f) Wen et al. \cite{wen2020simple}, (g) Lv et al. \cite{lv2021blind}, (h) Lv* et al. \cite{lv2021blind}, (i) Ours, (j) Ours*.}
\end{figure}
\begin{figure}
  \centering
  \subfigure[\scriptsize{Ground truth}]{\includegraphics[scale=0.33]{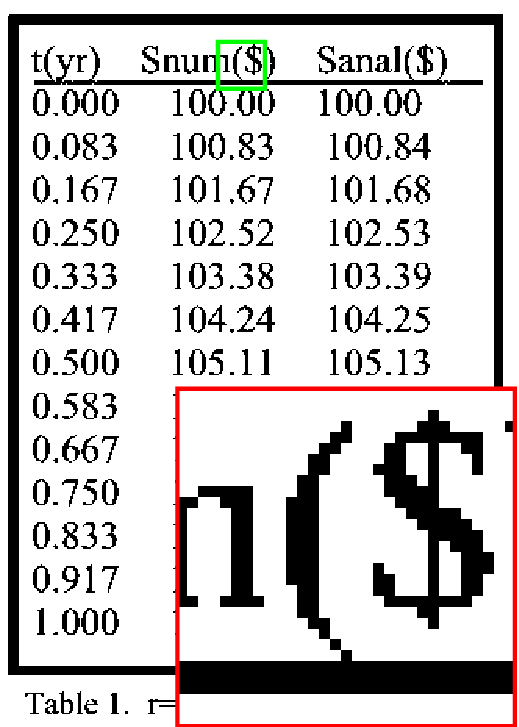}}
  \subfigure[\scriptsize{Blurry image}]{\includegraphics[scale=0.33]{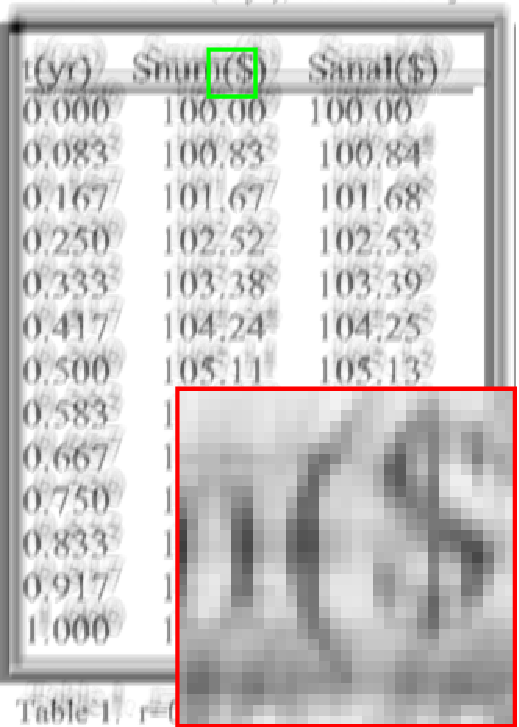}}
  \subfigure[\scriptsize{22.19/0.9567}]{\includegraphics[scale=0.33]{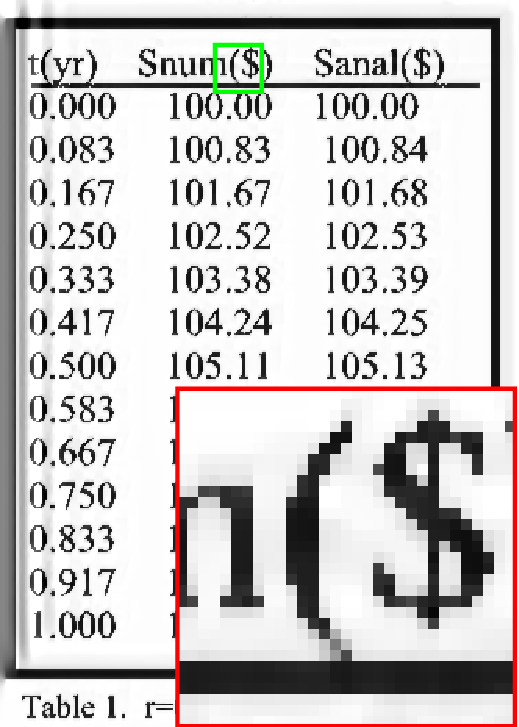}}
  \subfigure[\scriptsize{21.35/0.9449}]{\includegraphics[scale=0.33]{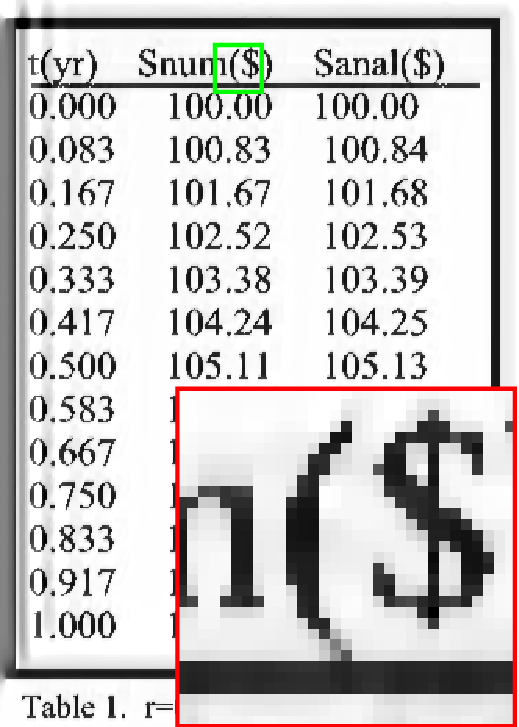}}
  \subfigure[\scriptsize{33.58/0.9984}]{\includegraphics[scale=0.33]{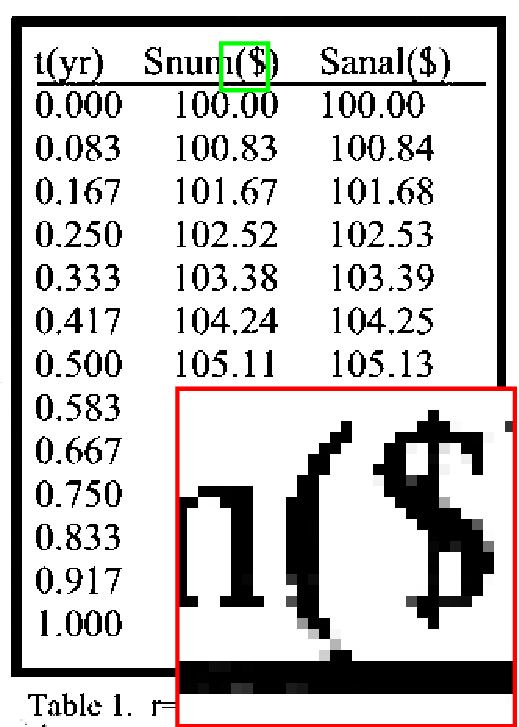}}
  \subfigure[\scriptsize{22.10/0.9570}]{\includegraphics[scale=0.33]{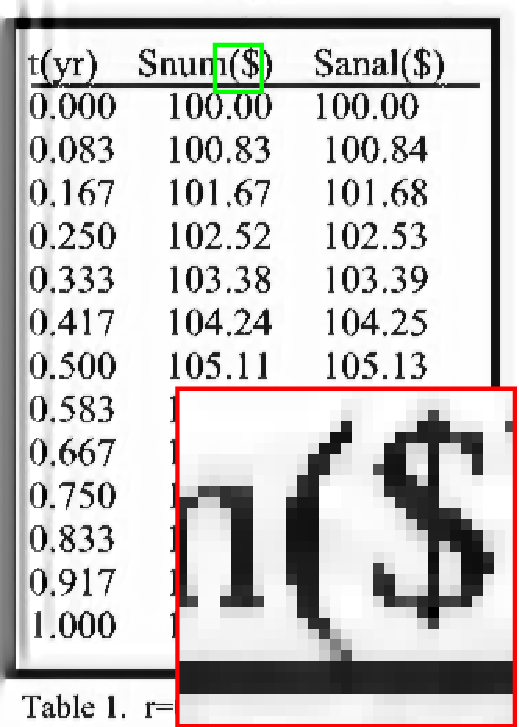}}
  \subfigure[\scriptsize{22.29/0.9586}]{\includegraphics[scale=0.33]{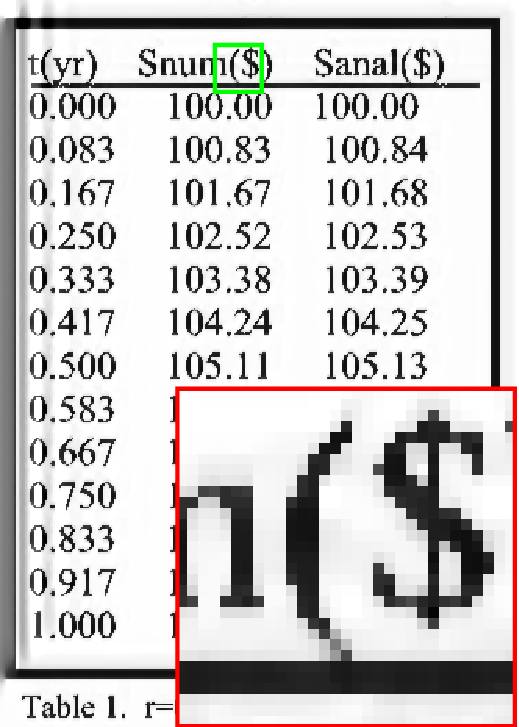}}
  \subfigure[\scriptsize{22.79/0.9604}]{\includegraphics[scale=0.33]{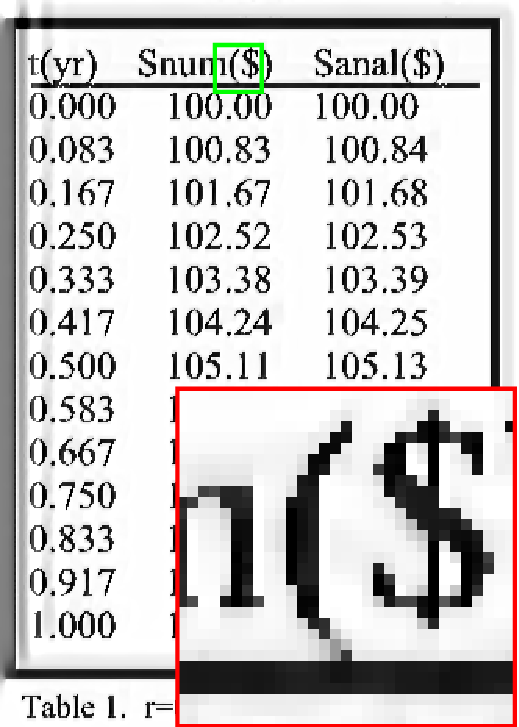}}
  \subfigure[\scriptsize{42.84/0,9995}]{\includegraphics[scale=0.33]{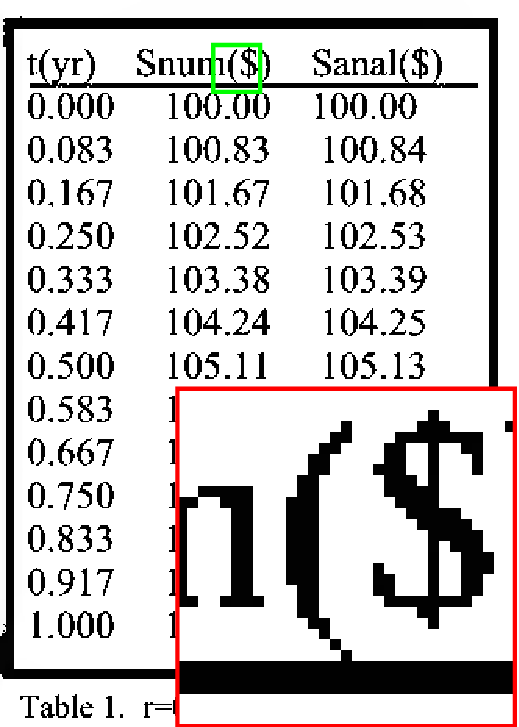}}
  \subfigure[\scriptsize{43.16/0.9996}]{\includegraphics[scale=0.33]{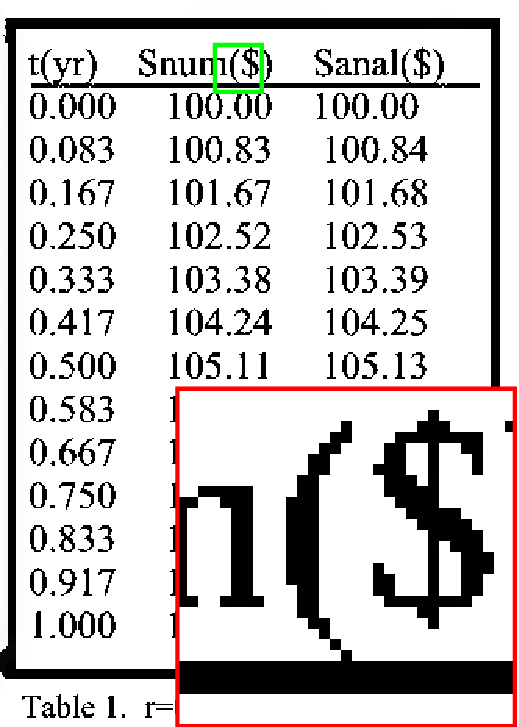}}
  \caption{The visual quality and quantitative results (PSNR/SSIM)  of different methods on ``im6''. (a) Ground truth, (b) Blurred image, (c) Pan et al. \cite{pan2016l_0},  (d) Chen et al. \cite{chen2019blind}, (e) Ren et al. \cite{ren2020neural}, (f) Wen et al. \cite{wen2020simple}, (g) Lv et al. \cite{lv2021blind}, (h) Lv* et al. \cite{lv2021blind}, (i) Ours, (j) Ours*.}
\end{figure}

\subsection{Blind pattern image deconvolution}

%%Pattern images are a significant category of images commonly encountered in human-made objects, artistic designs, and paintings. An important characteristic of pattern images is that their visual contents are depicted using a limited number of distinct pixel intensities \cite{mei2015unihist}.
On the second synthetic dataset consisting of 32 blurry pattern images, we compare the recovered results against several competing algorithms \cite{pan2016l_0,chen2019blind,ren2020neural,wen2020simple,lv2021blind}.
According to the results presented in Table 2 and Figure 9, it can be seen that our method yields a substantial improvement in the average PSNR, with an increase of at least 3.97 dB, and the average SSIM, with an improvement of at least 0.0126.
Figure 10 and Figure 11 display the visual comparison of deblurring results for two pattern images ``im10'' and ``im12'', respectively.
Compared to the competing methods, it can be observed that our proposed approach significantly enhances the visual quality with much less artifacts and sharper edges.
\begin{figure}
  \centering
  \subfigure[Average PSNR values]{\includegraphics[scale=0.38]{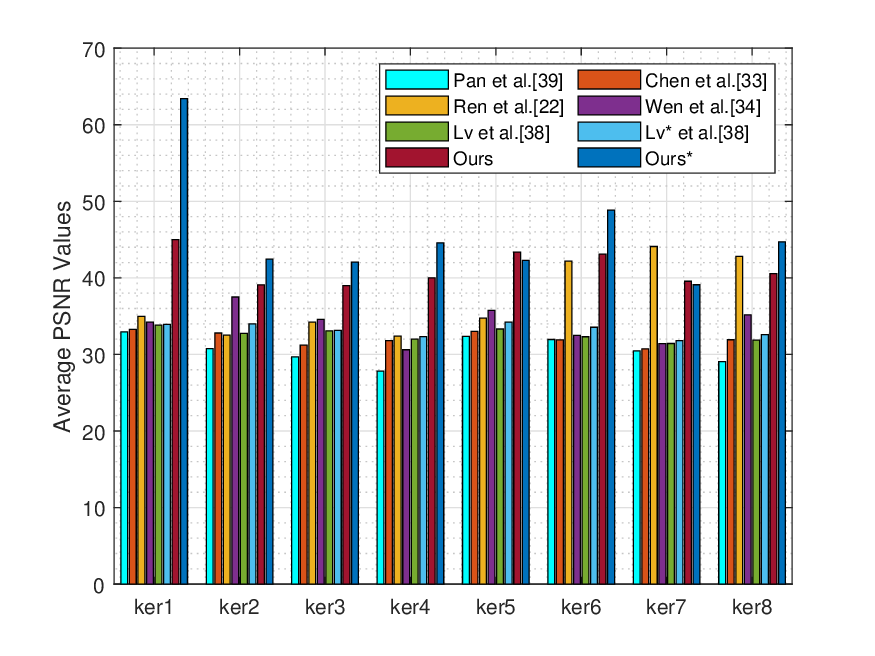}}
  \subfigure[Average SSIM values]{\includegraphics[scale=0.38]{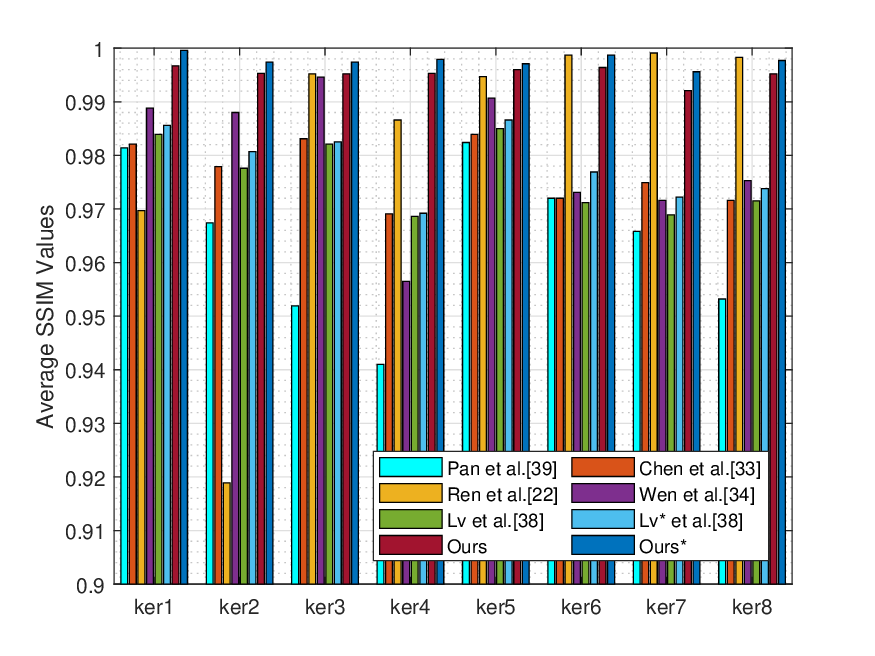}}
  \caption{Quantitative evaluations on 4 pattern images of different blur kernels. It can be observed that our method performs comparably against other competing methods. }
\end{figure}

\begin{table}
\centering
\caption{Quantitative results (average PSNR/SSIM) of pattern image deblurring.}
\footnotesize
\setlength{\tabcolsep}{11mm}{
\begin{tabular}{ccc}
\toprule
Method&PSNR $\uparrow$ &SSIM $\uparrow$ \\
\midrule
Pan et al. \cite{pan2016l_0}&30.64&0.9644\\
Chen et al. \cite{chen2019blind}&32.08&0.9768\\
Ren et al. \cite{ren2020neural}&37.24&0.9827\\
Wen et al. \cite{wen2020simple}&33.97&0.9798\\
Lv et al. \cite{lv2021blind}&32.57&0.9761\\
Lv* et al. \cite{lv2021blind}&33.20&0.9784\\
Ours&41.21&0.9953\\
Ours*&45.93&0.9977\\
\bottomrule
\end{tabular}}
\end{table}

\begin{figure}
  \centering
  \subfigure[\scriptsize{Ground truth}]{\includegraphics[scale=0.25]{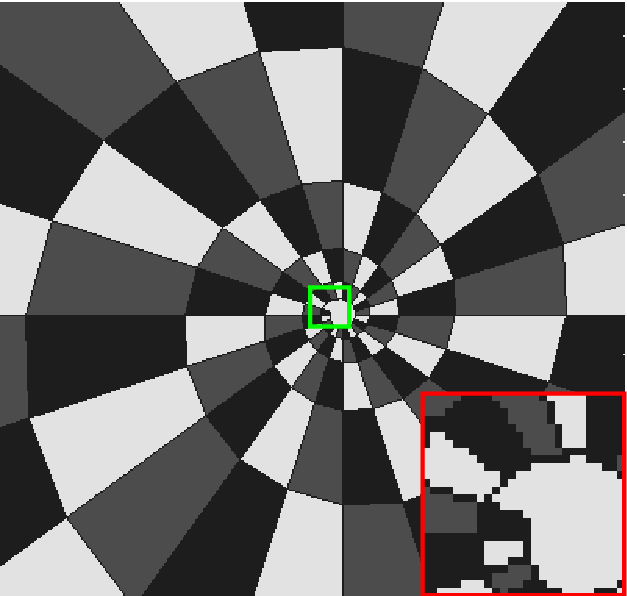}}
  \subfigure[\scriptsize{Blurry image}]{\includegraphics[scale=0.25]{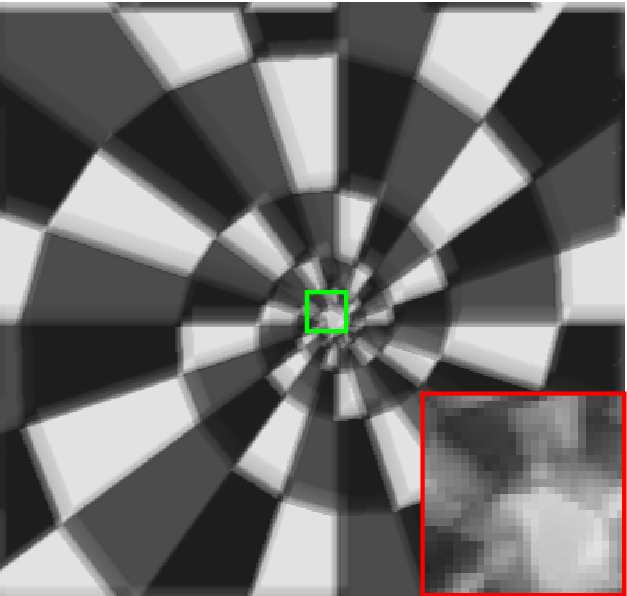}}
  \subfigure[\scriptsize{30.35/0.9757}]{\includegraphics[scale=0.25]{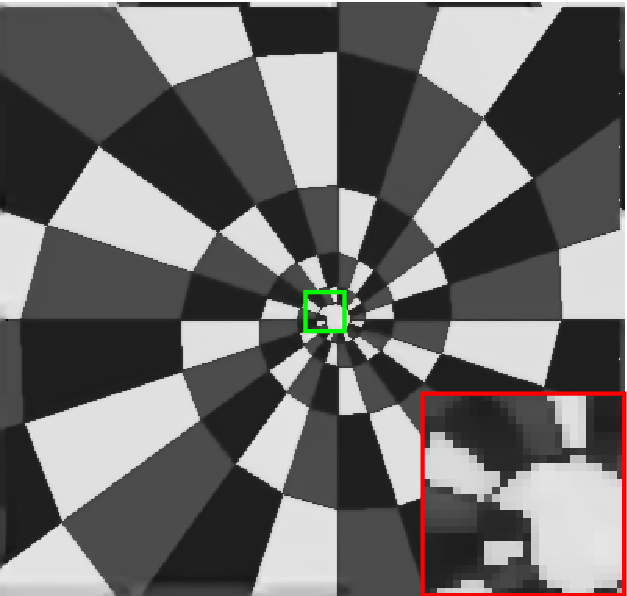}}
  \subfigure[\scriptsize{28.07/0.9590}]{\includegraphics[scale=0.25]{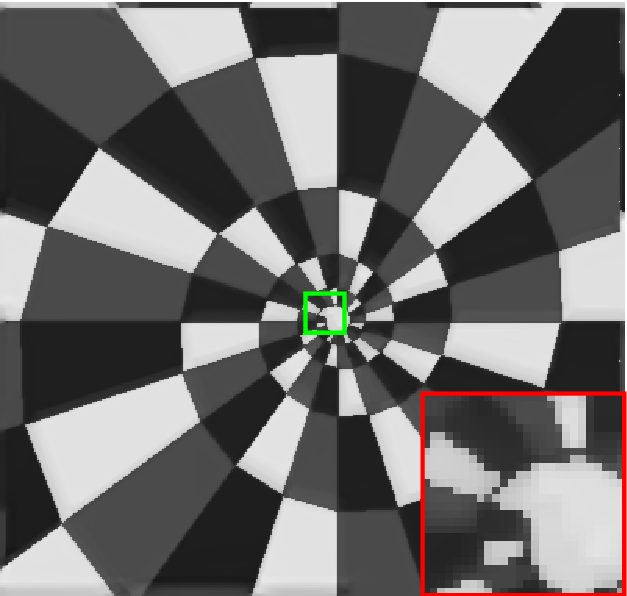}}
  \subfigure[\scriptsize{35.67/0.9966}]{\includegraphics[scale=0.25]{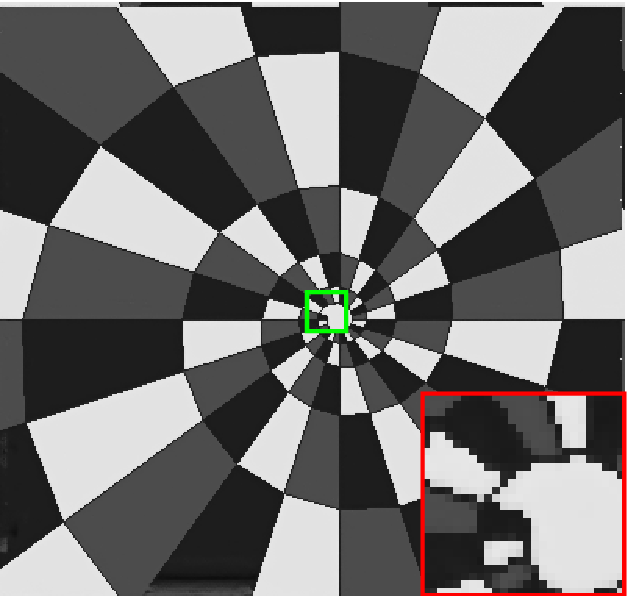}}
  \subfigure[\scriptsize{26.56/0.9520}]{\includegraphics[scale=0.25]{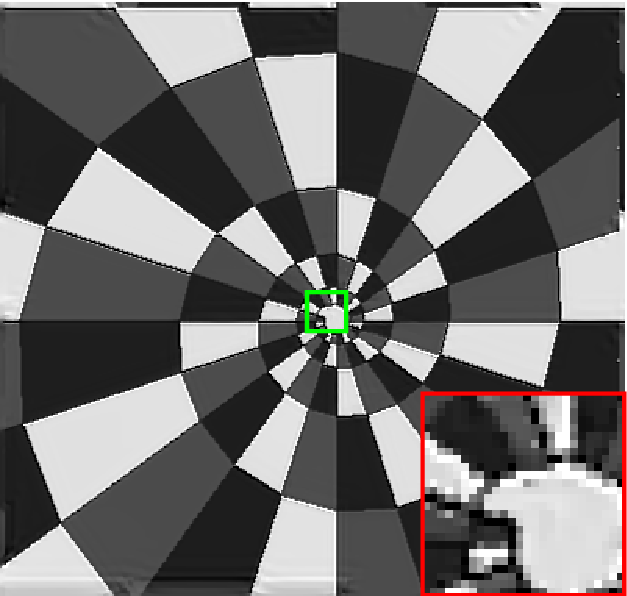}}
  \subfigure[\scriptsize{28.06/0.9588}]{\includegraphics[scale=0.25]{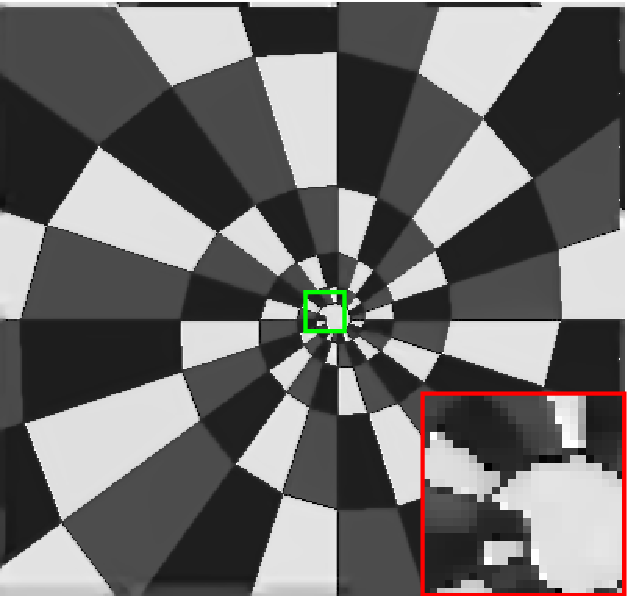}}
  \subfigure[\scriptsize{30.56/0.9762}]{\includegraphics[scale=0.25]{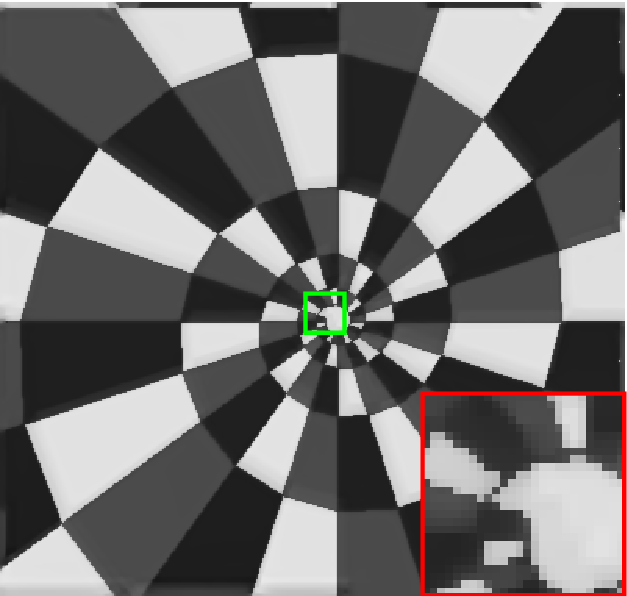}}
  \subfigure[\scriptsize{38.37/0.9976}]{\includegraphics[scale=0.25]{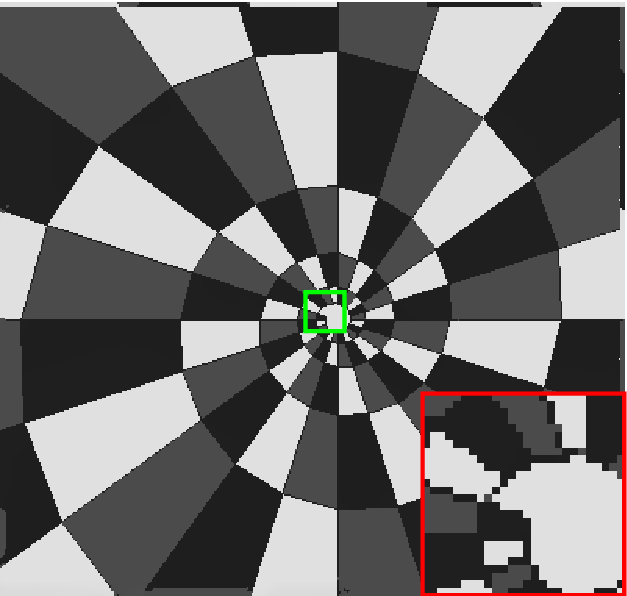}}
  \subfigure[\scriptsize{39.15/0.9980}]{\includegraphics[scale=0.25]{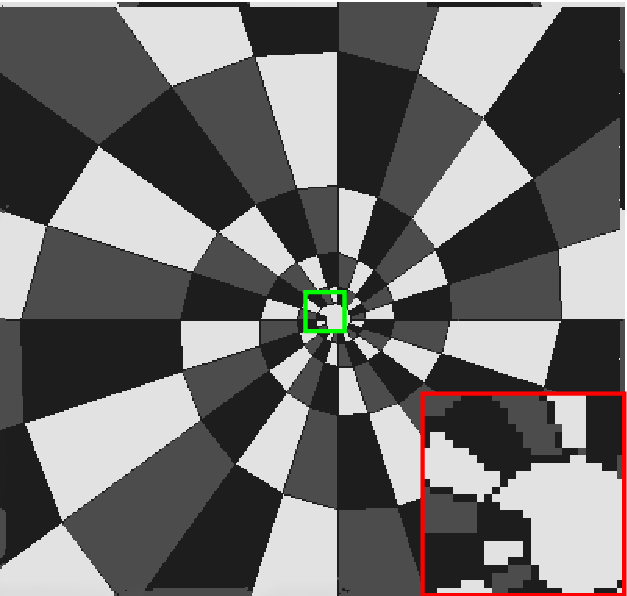}}
  \caption{The visual quality and quantitative results (PSNR/SSIM)  of different methods on ``im10''. (a) Ground truth, (b) Blurred image, (c) Pan et al. \cite{pan2016l_0},  (d) Chen et al. \cite{chen2019blind}, (e) Ren et al. \cite{ren2020neural}, (f) Wen et al. \cite{wen2020simple}, (g) Lv et al. \cite{lv2021blind}, (h) Lv* et al. \cite{lv2021blind}, (i) Ours, (j) Ours*.}
\end{figure}
\begin{figure}
  \centering
  \subfigure[\scriptsize{Ground truth}]{\includegraphics[scale=0.24]{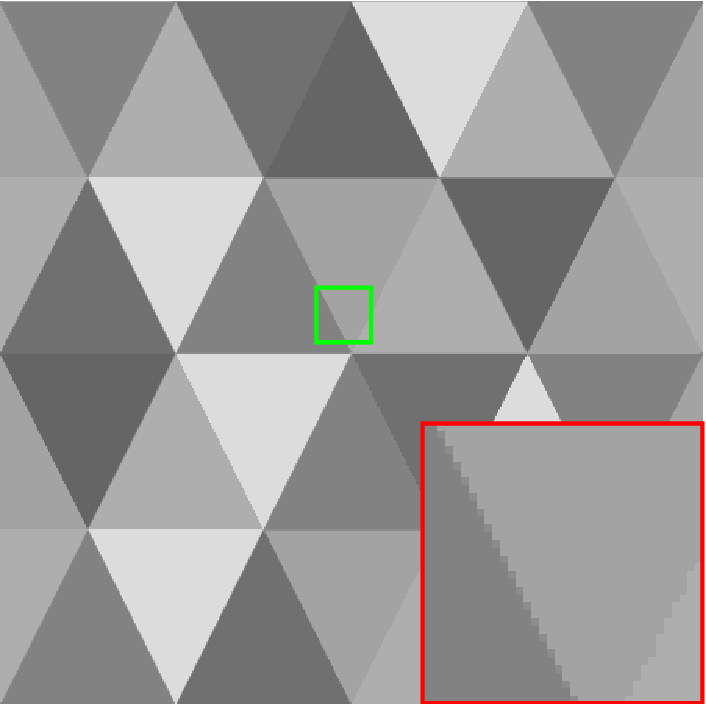}}
  \subfigure[\scriptsize{Blurry image}]{\includegraphics[scale=0.24]{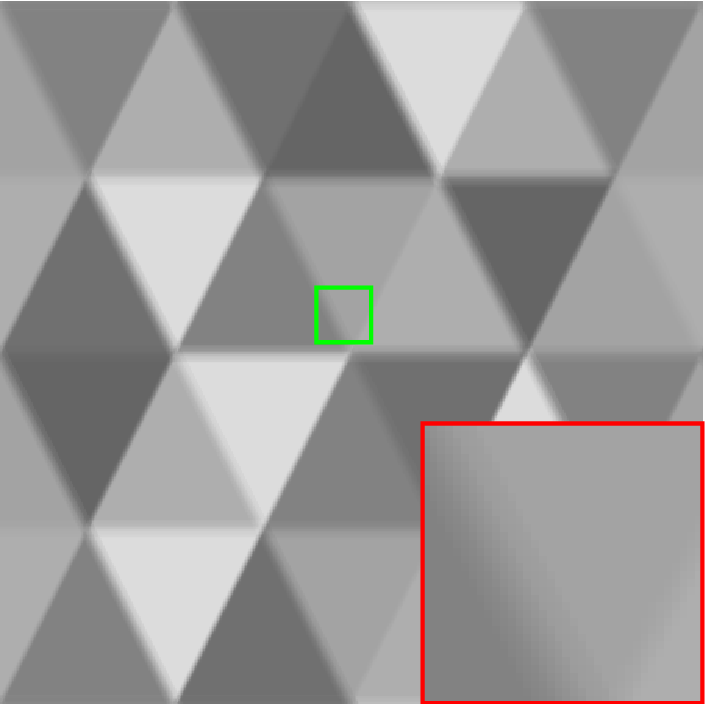}}
  \subfigure[\scriptsize{34.91/0.9839}]{\includegraphics[scale=0.24]{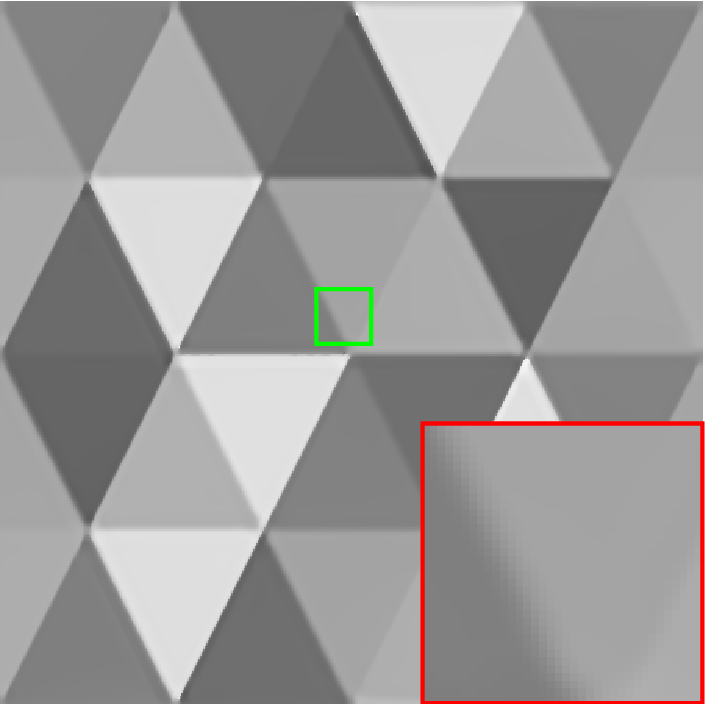}}
  \subfigure[\scriptsize{36.54/0.9894}]{\includegraphics[scale=0.24]{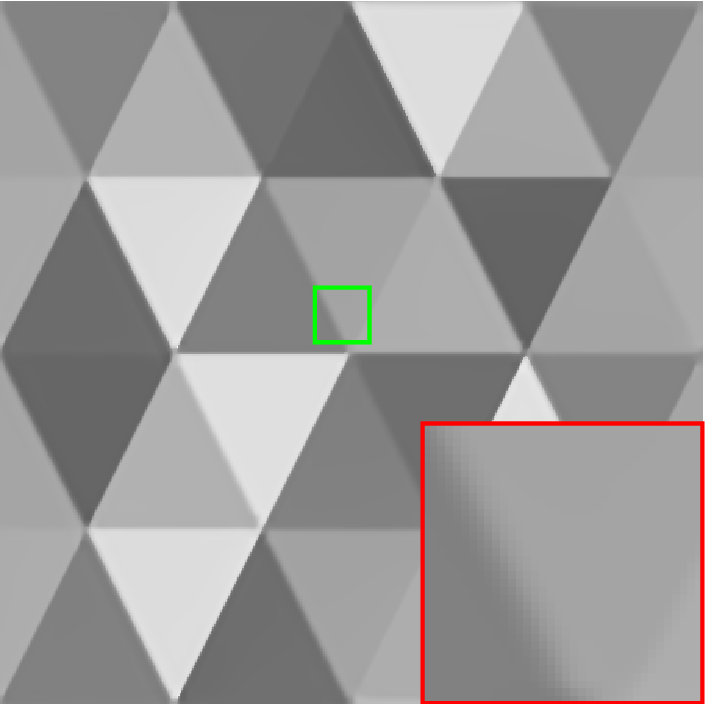}}
  \subfigure[\scriptsize{40.79/0.9982}]{\includegraphics[scale=0.24]{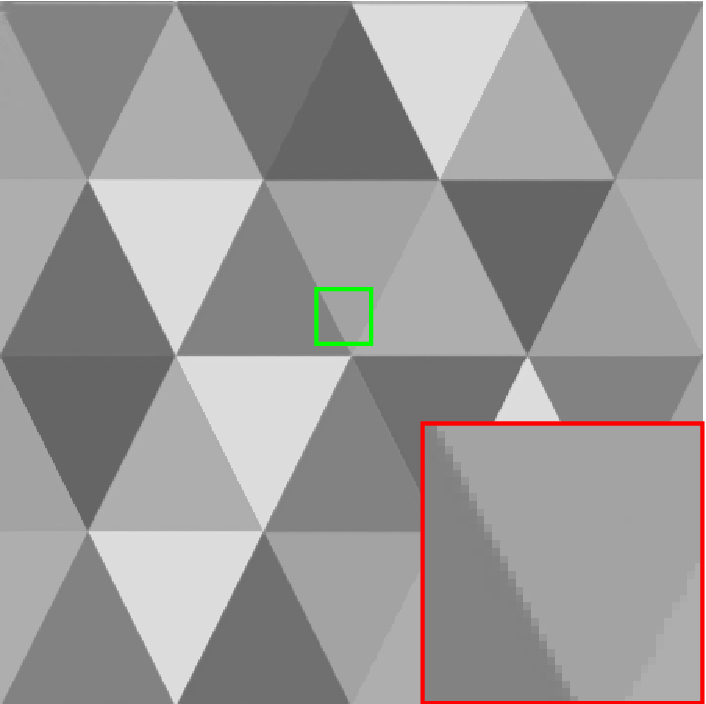}}
  \subfigure[\scriptsize{37.41/0.9943}]{\includegraphics[scale=0.24]{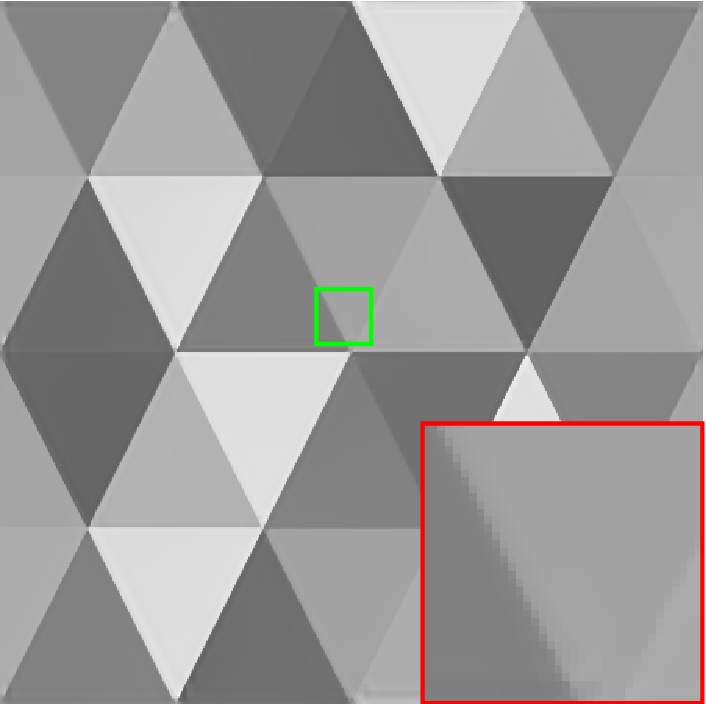}}
  \subfigure[\scriptsize{36.77/0.9893}]{\includegraphics[scale=0.24]{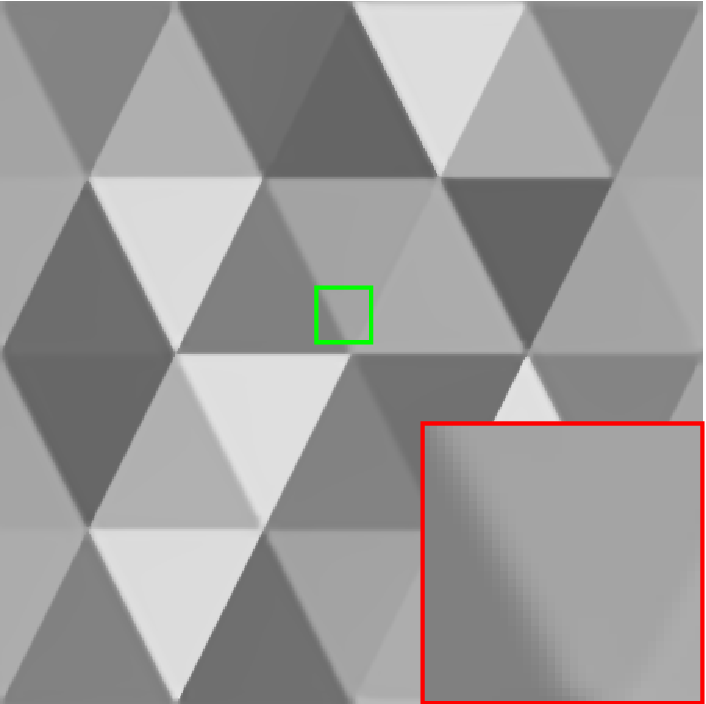}}
  \subfigure[\scriptsize{37.08/0.9903}]{\includegraphics[scale=0.24]{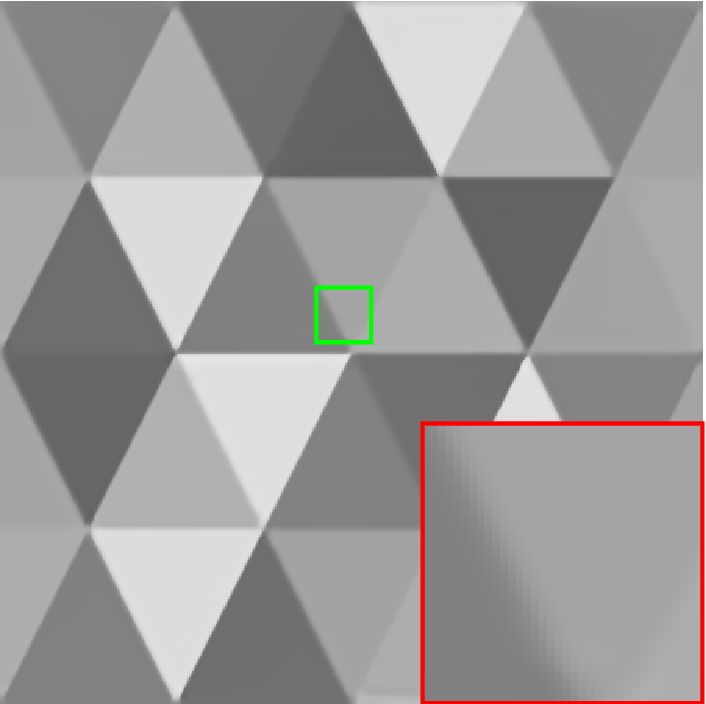}}
  \subfigure[\scriptsize{41.30/0.9970}]{\includegraphics[scale=0.24]{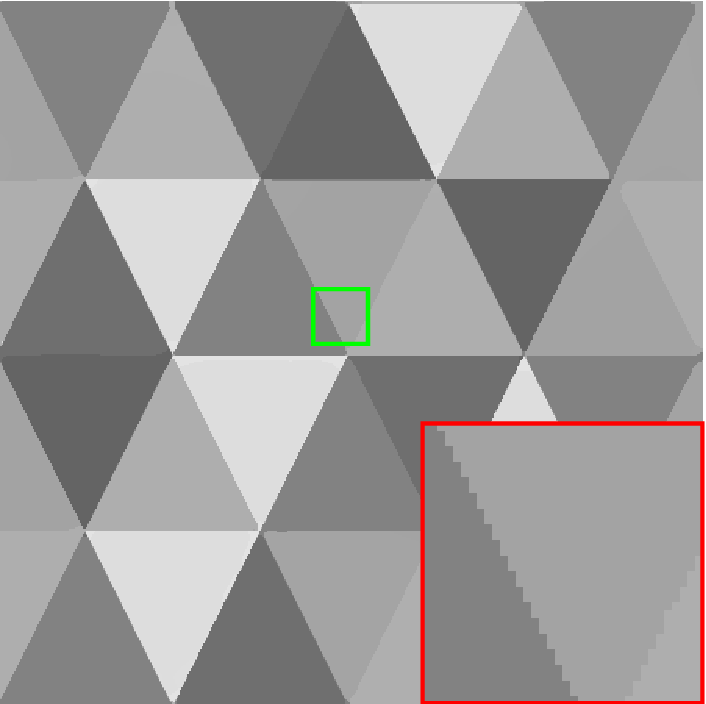}}
  \subfigure[\scriptsize{42.40/0.9996}]{\includegraphics[scale=0.24]{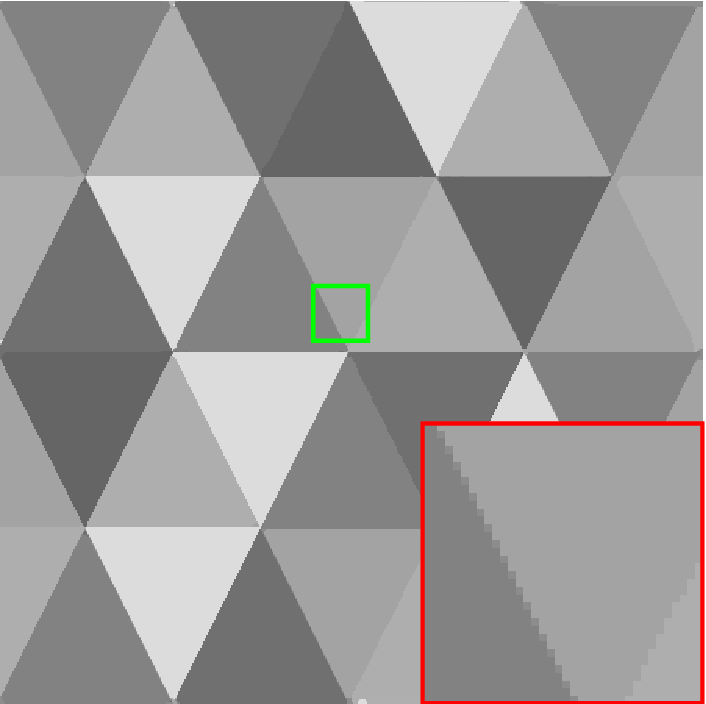}}
  \caption{The visual quality and quantitative results (PSNR/SSIM)  of different methods on ``im12''. (a) Ground truth, (b) Blurred image, (c) Pan et al. \cite{pan2016l_0},  (d) Chen et al. \cite{chen2019blind}, (e) Ren et al. \cite{ren2020neural}, (f) Wen et al. \cite{wen2020simple}, (g) Lv et al. \cite{lv2021blind}, (h) Lv* et al. \cite{lv2021blind}, (i) Ours, (j) Ours*.}
\end{figure}

\subsection{Blurry and noisy binary image deconvolution}

On the third synthetic dataset, we evaluate the proposed method for blurry and noisy binary image deconvolution. As shown in Table 3 (BSNR = 40 dB) and Table 4 (BSNR = 30 dB), our method consistently achieves superior reconstruction performance across different images and blur kernels. At BSNR = 40 dB, our method obtains the highest PSNR for all four test images and achieves an SSIM of 0.9999 for the first three images, substantially outperforming the competing BID methods. For example, compared with the best competing result of Lv* et al. \cite{lv2021blind}, our method improves the PSNR from 34.54 dB to 50.20 dB on im1 and from 32.41 dB to 74.38 dB on im2. These results demonstrate that explicitly incorporating the restricted pixel-intensity prior of binary images, together with gradient sparsity regularization, can effectively suppress the effects of blur and noise and recover binary structures with high fidelity. The experiments at BSNR = 30 dB further verify the robustness of the proposed method under stronger noise.

\begin{table}
\centering
\caption{Quantitative results (PSNR/SSIM) of blurry and noisy (BSNR = 40 dB) binary image deblurring.}
\footnotesize
\setlength{\tabcolsep}{3.4mm}{
\begin{tabular}{ccccc}
\toprule
Image (kernel) &im1 (ker1) &im2 (ker2) &im3 (ker3) &im4 (ker4)  \\
\midrule
Pan et al. \cite{pan2016l_0}&19.46/0.9250 &30.70/0.9629 &10.09/0.6687 &18.80/0.7715 \\
Chen et al. \cite{chen2019blind}&24.86/0.9566 &29.92/0.9131 &17.39/0.9383 &19.43/0.6768\\
Ren et al. \cite{ren2020neural}&20.97/0.9519 &29.28/0.8163 &19.03/0.9580 &21.75/0.7839 \\
Wen et al. \cite{wen2020simple}&11.72/0.7550 &45.75/0.9999 &27.61/0.9951 &20.27/0.9051 \\
Lv et al. \cite{lv2021blind}&32.69/0.9847 &28.82/0.9103 &15.76/0.8891 &21.90/0.7866 \\
Lv* et al. \cite{lv2021blind}&34.54/0.9930 &32.41/0.9914 &20.06/0.9702 &22.78/0.8653 \\
Ours&50.20/0.9999 &74.38/0.9999 &50.98/0.9999 &30.61/0.7703 \\
Ours*&74.43/0.9999 &79.65/0.9999 &78.18/0.9999 &48.90/0.9987 \\
\bottomrule
\end{tabular}}
\end{table}

\begin{table}
\centering
\caption{Quantitative results (PSNR/SSIM) of blurry and noisy  (BSNR = 30 dB)  binary  image deblurring.}
\footnotesize
\setlength{\tabcolsep}{3.4mm}{
\begin{tabular}{ccccc}
\toprule
Image (kernel)  &im5 (ker5) &im6 (ker6) &im7 (ker7) &im8 (ker8) \\
\midrule
Pan et al. \cite{pan2016l_0} &31.70/0.8436 &24.33/0.9726 &23.37/0.9422 &32.63/0.9919  \\
Chen et al. \cite{chen2019blind} &29.86/0.8926 &23.88/0.9378 &22.25/0.9169 &26.66/0.9582  \\
Ren et al. \cite{ren2020neural} &31.80/0.9284 &25.91/0.9836 &23.33/0.9458 &31.74/0.9904 \\
Wen et al. \cite{wen2020simple} &40.63/0.9325 &28.96/0.9929 &24.40/0.9673 &27.39/0.9885 \\
Lv et al. \cite{lv2021blind} &31.57/0.9273 &26.53/0.9858 &23.62/0.9469 &28.16/0.9876 \\
Lv* et al. \cite{lv2021blind} &33.27/0.9776 &28.09/0.9895 &24.21/0.9488 &28.68/0.9689 \\
Ours &46.77/0.9777 &53.23/0.9999 &33.36/0.9813 &46.72/0.9899  \\
Ours* &46.98/0.9969 &66.55/0.9999 &33.51/0.9799 &70.71/0.9999  \\
\bottomrule
\end{tabular}}
\end{table}

\subsection{Convergence analysis}
Our method utilizes the HQS technique to address the non-convex and constrained minimization problem (\ref{eq: proposed constraint1 model}).
Due to the involvement of multiple auxiliary variables during the optimization process, proving the overall convergence property theoretically is challenging.
Instead, following previous methods \cite{pan2016blind,pan2016l_0}, we empirically evaluate the convergence behavior of our algorithm by measuring the PSNR values and kernel similarity \cite{hu2012good}.
As shown in Figure 12, as the number of iterations increases, one can observe that both the PSNR curve and kernel similarity curve gradually increase and ultimately become flat and stable, demonstrating reliable convergence behavior of the proposed algorithm.
%%Our experiments were conducted at the finest image scale using a set of 12 images (im1-im12) that were blurred with ker1.
%%we empirically evaluate the convergence of the proposed method based on PSNR values of the restored results at the finest image scale.

\begin{figure}
  \centering
  \subfigure[PSNR value v.s. Iteration]{\includegraphics[scale=0.38]{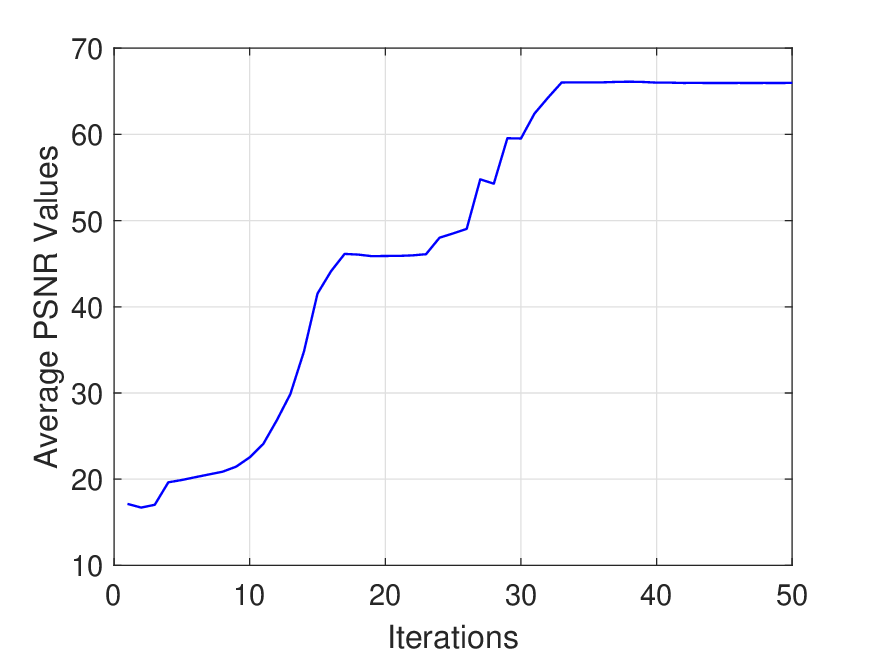}}
  \subfigure[Kernel similarity v.s. Iteration]{\includegraphics[scale=0.38]{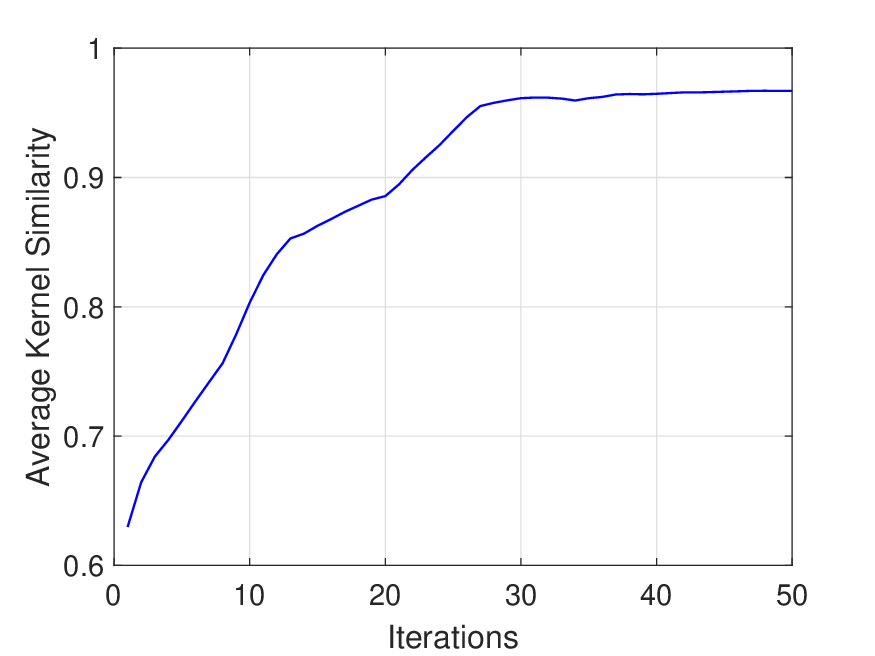}}
  \caption{The convergence behavior of our proposed algorithm. The test set contains 12 images (im1-im12) that are blurred with kernel ``ker1''. Note that the PSNR values and the kernel similarity are evaluated at the finest image scale.
  }
\end{figure}

\subsection{Effect of the number $s$}
Recall that the set $\Omega$ of distinct pixel intensities $\{\alpha_{1}, \alpha_{2}, \ldots, \alpha_{s}\}$ is obtained by applying the K-Means technique on the reference image obtained from an off-the-shelf BID approach. Thus, the effect of the number $s$ (i.e., clustering number in K-Means method) in the proposed model (4) is also a matter of significant concern. We display the influence of the parameter $s$ in two scenarios: ``im2'' (with the true index $s=2$) blurred by kernel ``ker2'', and ``im12'' (with the true index $s=7$) also blurred by kernel ``ker2'', while keeping other parameters fixed. Figure 13 illustrates the influence of $s$ on the algorithm's performance, with $s$ ranging from 2 to 5 and 5 to 9 (with a step size of 1), respectively.
As expected, it can be seen that the SSIM values reach their maximum under the ideal $s$ setting and gradually decrease as $s$ deviates from the true values. This indicates that the proposed method can deliver comparable performance within an appropriate range of the number $s$.
%This section discusses how to determine the parameter $s$ and how the performance of our algorithm is affected by the parameter $s$.

\begin{figure}
  \centering
  %\subfigure[Intensity histogram]{\includegraphics[scale=0.42]{spixel.eps}}
  \subfigure[SSIM value v.s. Number $s$]{\includegraphics[scale=0.38]{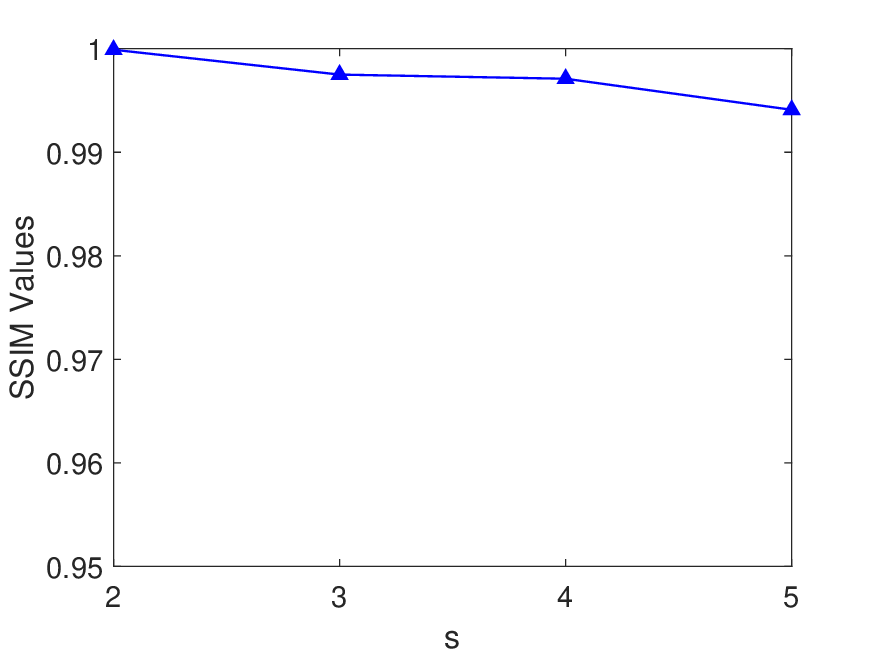}}
  \subfigure[SSIM value v.s. Number $s$]{\includegraphics[scale=0.38]{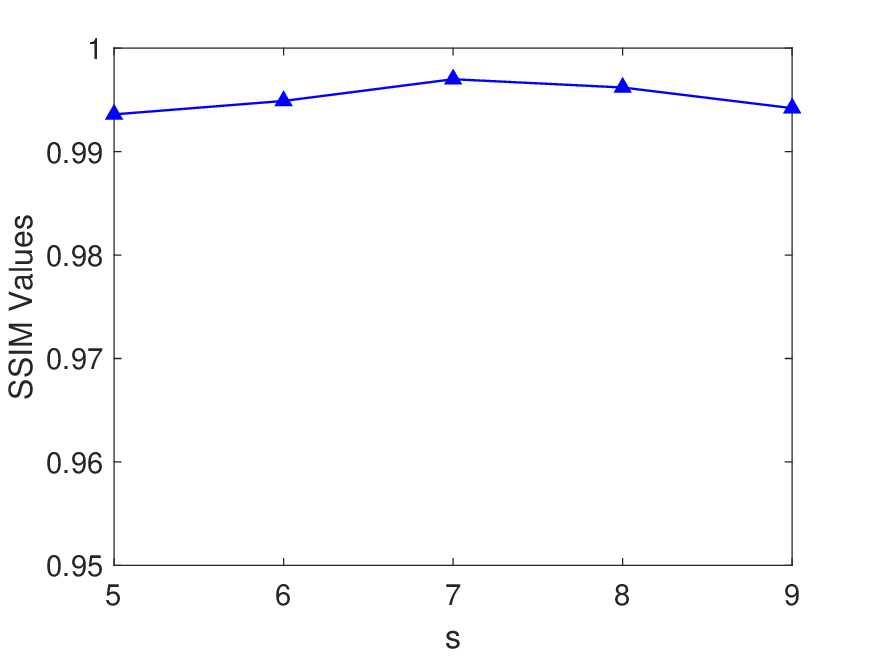}}
  \caption{The evolution of SSIM values with the parameter $s$. (a) Test image: ``im2'' (with an ideal value
  $s=2$), blur kernel: ker2. (b) Test image: ``im12'' (with an ideal value $s=7$), blur kernel: ker2.}
%  \caption{ The sensibility of parameter $s$ on the multi-valued pattern image ``im11'' (with an ideal value $s=3$), blur kernel: ker6. (a) Evolution of SSIM values with the parameter $s$.}
  %(a) The pixel intensity histogram of the reference image,

\end{figure}

\subsection{Running time}
Table 5 presents the running times of our algorithm and competing approaches \cite{pan2016l_0,chen2019blind,wen2020simple,lv2021blind} on 24 blurry images with 3 different resolutions. It can be observed that our method achieves highly competitive efficiency compared to other competing methods. Note that the running time for the non-blind deblurring implementation is not included in all these methods.
%%In Table 3, it can be observed that our method requires a shorter running time than most competing methods (note that the running time for the non-blind deblurring implementation is not included in the results for both our approach and the comparison methods).

\begin{table}[htbp]
\centering
\caption{Average running time (in seconds) comparisons. The kernel size is $35\times35$.}
\footnotesize
\setlength{\tabcolsep}{6.5mm}{
\begin{tabular}{cccc}
\toprule
Image size &$256\times256$&$450\times450$&$600\times600$\\
\midrule
Pan et al. \cite{pan2016l_0}&13.28&42.07&94.23\\
Chen et al. \cite{chen2019blind}&53.66&147.87&288.39\\
Wen et al. \cite{wen2020simple}&12.60&20.37&31.33\\
Lv et al. \cite{lv2021blind}&16.48&45.26&83.26\\
Ours&10.38&32.66&59.83\\
\bottomrule
\end{tabular}}
\end{table}

\section{Analysis and discussion}

\subsection{Advantages}
The proposed algorithm imposes a strict constraint on the pixel intensities of the recovered image, fully exploiting the inherent nature of binary and pattern images, where pixel intensities can only take a very limited number of values. As demonstrated in Figure 14, the proposed method preserves the distribution characteristics of pixel intensities and effectively suppresses unappealing artifacts. By integrating the pixel intensity constraint into the gradient-sparsity regularizer model, as shown in Figure 15, our method can eliminate undesirable structures more effectively in the final result compared to methods relying solely on sparse gradient priors. Moreover, our algorithm does not depend on additional non-blind deblurring approaches and can directly produce satisfactory results. It is important to note that employing a two-stage strategy to restore final results with an additional implementation of a non-blind deconvolution approach may potentially reduce the accuracy of the pixel distribution in the restored image, resulting in undesirable artifacts, as shown in Figure 15(d).
%Moreover, our method combines both the advantages of the above pixel intensity prior and the sparse gradient prior, as shown in Figure 14, more productively eliminating undesirable structures in the final result compared to the method relying solely on the sparse gradient prior.

\begin{figure}[htbp]
  \centering
  \subfigure[]{\raisebox{0.6cm}{\includegraphics[scale=0.26]{kblur2.eps}}}
  \subfigure[]{\includegraphics[scale=0.26]{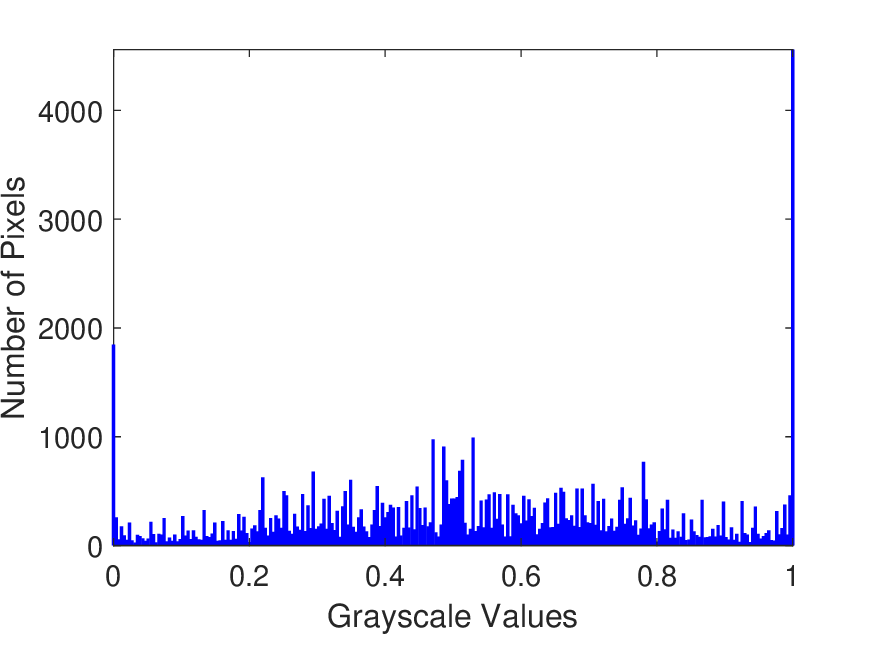}}
  \subfigure[]{\raisebox{0.6cm}{\includegraphics[scale=0.26]{kpan2.eps}}}
  \subfigure[]{\includegraphics[scale=0.26]{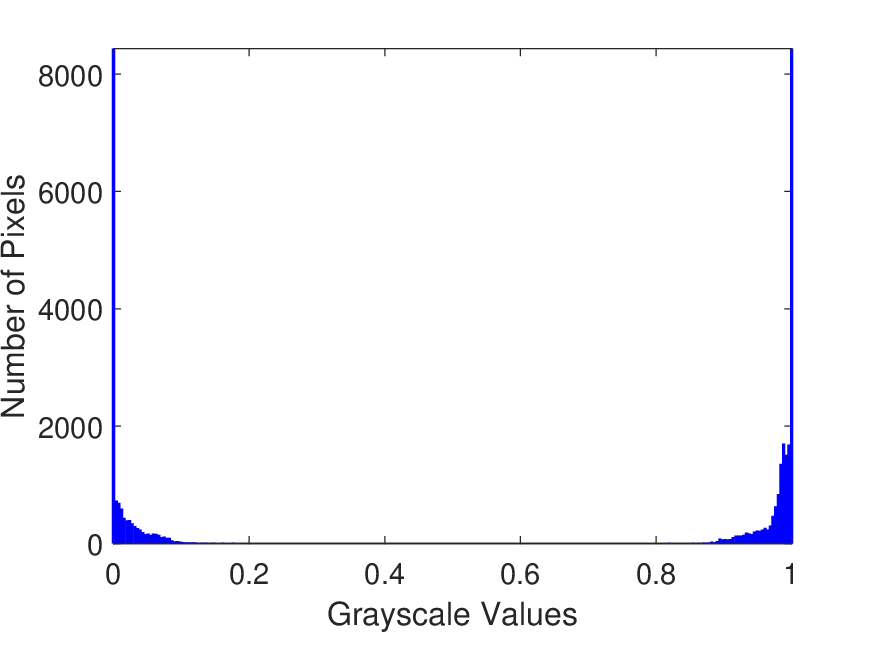}}\\
  \subfigure[]{\raisebox{0.6cm}{\includegraphics[scale=0.26]{klv2.eps}}}
  \subfigure[]{\includegraphics[scale=0.26]{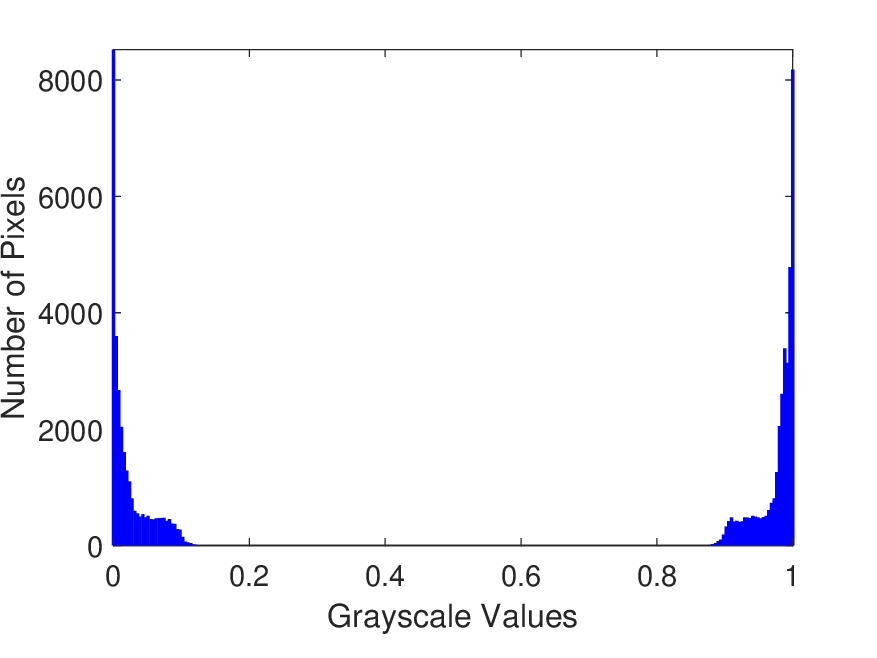}}
  \subfigure[]{\raisebox{0.6cm}{\includegraphics[scale=0.26]{kour2.eps}}}
  \subfigure[]{\includegraphics[scale=0.26]{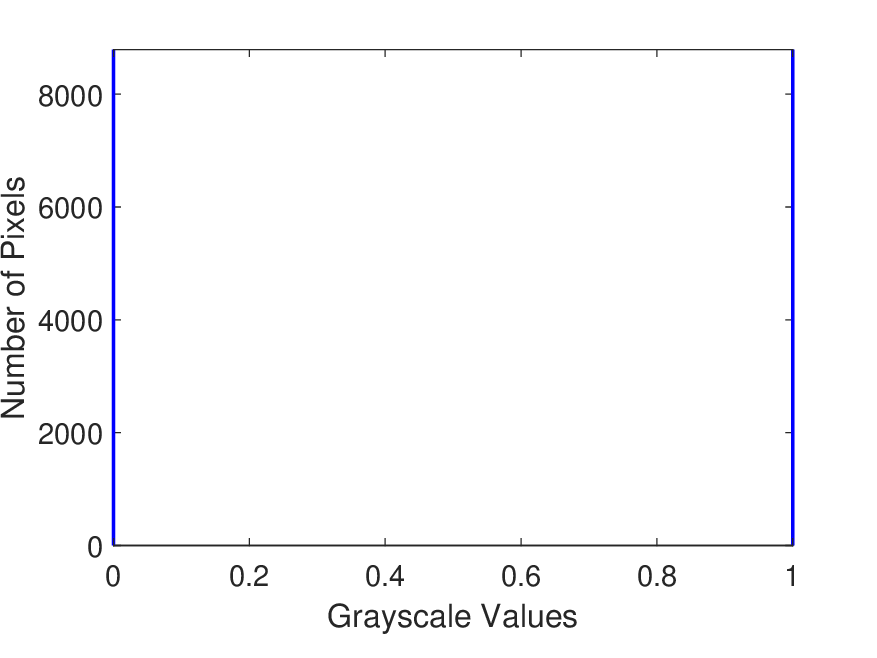}}
  \caption{The statistical property comparison between the blurry image and recovery results. (a) Blurred image and (b) the histograms of its intensity, (c) Recovery result of Pan et al. \cite{pan2016l_0} and (d) the histogram of its intensity, (e) Recovery result of Lv* et al. \cite{lv2021blind} and (f) the histogram of its intensity, (g) Recovery result of ours* and (h) the histogram of its intensity. }
\label{Fig:12}
\end{figure}

\begin{figure}[htbp]
  \centering
  \subfigure[\scriptsize{Input}]{\includegraphics[scale=0.41]{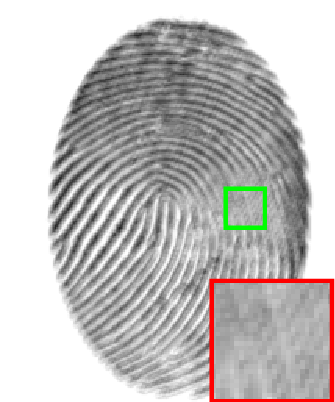}}
  \subfigure[\scriptsize{8.18/0.3600}]{\includegraphics[scale=0.41]{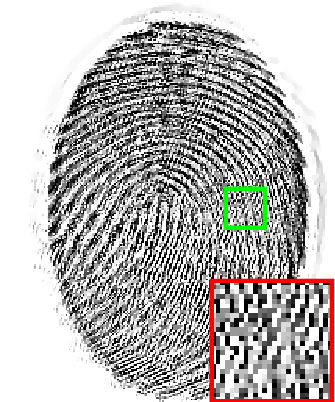}}
  \subfigure[\scriptsize{10.97/0.7379}]{\includegraphics[scale=0.41]{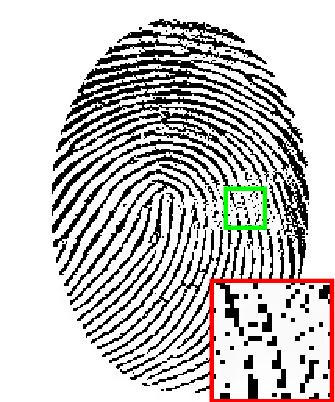}}
  \subfigure[\scriptsize{22.42/0.9812}]{\includegraphics[scale=0.41]{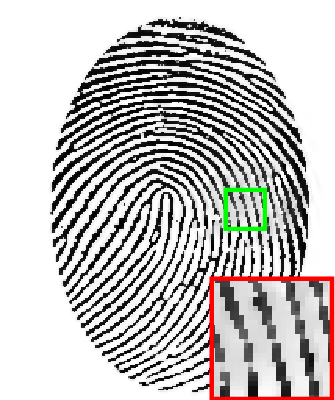}}
  \subfigure[\scriptsize{77.96/0.9999}]{\includegraphics[scale=0.41]{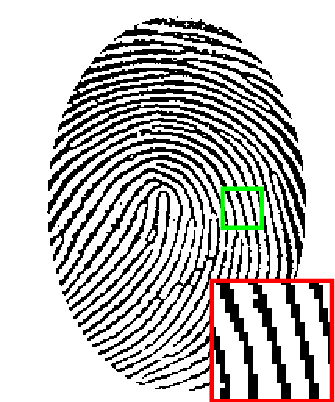}}
  \caption{Ablation study on binary image ``im3'' (PSNR/SSIM). (a) Input, (b) Ours* without pixel intensity constraint, (c) Ours* without gradient sparsity prior, (d) Ours* with non-blind deblurring algorithm \cite{lv2021blind}, (e) Ours*.}
\end{figure}
%the estimated binary and pattern images achieve significant improvement in both visual quality and quantitative measures. It is not mandatory for our algorithm to utilize another non-blind deconvolution method to generate the final deblurring result.

%and it is not a compulsory choice for our algorithm to adopt another non-blind deconvolution method to generate final deblurring result.
%achieve favorable numerical and visual representation
%In contrast, our method simultaneously estimates both the blur kernel and the sharp image, which achieves the state-of-the-art performance, clearly demonstrated in Figures 5-8.

\subsection{Limitations and possible improvements}
Although our proposed method has been empirically demonstrated to yield promising BID results for two-valued binary and multi-valued pattern images, it has its own applicable conditions and limitations, because the accuracy of the pixel intensity set
$\Omega$ significantly affects the performance of our method. Specifically, we will elaborate on two aspects: the choice of the reference image and the strategy for estimating pixel intensities.

It is noted that the proposed algorithm prefers high-quality reference images $\mathbf{x}^{\mathrm{ref}}$ over low-quality ones. While the algorithm can tolerate a certain amount of false detections of
$\Omega$, the error cannot be too large. When the quality of the reference image is quite poor (e.g., the input blurry image serves as $\mathbf{x}^{\mathrm{ref}}$), the detected support set might be completely unreliable, resulting in very low-quality recovery results. From another perspective, since the initial reference image can be provided by any off-the-shelf BID method, we have high flexibility in choosing a high-quality reference image.

Additionally, after obtaining the reference image
$\mathbf{x}^{\mathrm{ref}}$, we only applied a simple K-Means clustering method to approximate the pixel intensities in this preliminary work. Table 4 shows a comparison between the estimated pixel intensities by the K-Means method and the ideal true values. Although the values are relatively close, there is still a slight discrepancy. This also indicates that there is room for improvement in the proposed algorithm by leveraging a more sophisticated pixel intensity estimation strategy and enhancing the accuracy of the estimated set $\Omega$ in practical applications.

%it is subject to certain applicable conditions and limitations, that is, whether the pixel intensities set $\Omega=\{\alpha_{1},\alpha_{2},\cdot\cdot\cdot,\alpha_{s}\}$ could be accurately obtained significantly influence the recovery performance, thereby it would be reasonable to infer that if the target multi-valued pattern image contains a large number of distinct pixel values (\textit{i.e.,} the $s$ is large), the performance of proposed method may be weaken and limited due to the interference between adjacent pixel values.
%%In this preliminary work, we only adopted a simple K-Means clustering method, sometimes leading to inexact centers of cluster as the image degradation process blurs across distinct pixel values and shrinks the distances between them.
%%$\Omega=\{\alpha_{1},\alpha_{2},\cdot\cdot\cdot,\alpha_{s}\}$.

%%\begin{table}[htbp]
%%\centering
%%\caption{ The comparison between distinct pixel values estimated by K-Means method and the ground truth (blur kernel: ker1).}
%%\footnotesize
%%\setlength{\tabcolsep}{2mm}{
%%\begin{tabular}{|c||c|c|c|}
%%\hline
%%Test image &im4&im7&im10\\
%%\hline
%%True pixel intensities & 0 and 1 & 0 and 1 & 0.1137,0.2980,0.8863 \\
%%\hline
%%K-Means estimation & 0.0063 and 0.9945 & 0.0073 and 0.9970 & 0.1209,0.2910,0.8782 \\
%%\hline
%%\end{tabular}}
%%\end{table}

\begin{table}[htbp]
\centering
\caption{The comparison between distinct pixel values estimated by K-Means method and the ground truth (blur kernel: ker1). }
\footnotesize
\setlength{\tabcolsep}{2.5mm}{
\begin{tabular}{cccc}
\toprule
Test image &im4 &im7 &im10\\
\midrule
True pixel intensities & 0 and 1 & 0 and 1 & 0.1137,0.2980,0.8863 \\
K-Means estimation & 0.0063 and 0.9945 & 0.0073 and 0.9970 & 0.1209,0.2910,0.8782 \\

\bottomrule
\end{tabular}}
\end{table}

\subsection{Extension to other methods}
To illustrate the flexibility and generalizability of the proposed pixel intensity constraint, we extend it to the prevailing BID algorithm \cite{chen2019blind}. As demonstrated in Figure 16, the pixel intensity constraint compensates for the inability of \cite{chen2019blind} to handle binary images with rich texture details, resulting in superior visual and quantitative results. Similarly, when applied to complicated pattern images, the incorporation of our pixel intensity constraint effectively eliminates out-of-range pixel values, yielding visually pleasing images with significantly fewer ringing artifacts and sharper edges.

\begin{figure}[!h]
  \centering
  \subfigure[\scriptsize{Ground truth}]{\includegraphics[scale=0.52]{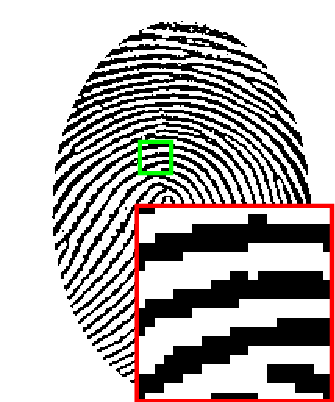}}
  \subfigure[\scriptsize{Blurred image}]{\includegraphics[scale=0.52]{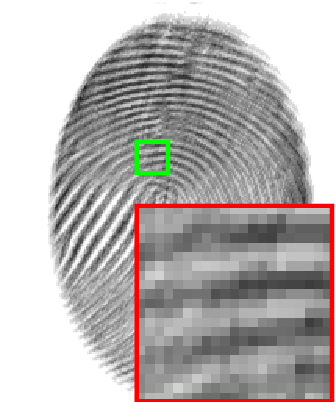}}
%  \subfigure[\scriptsize{15.89/0.9203}]{\includegraphics[scale=0.5]{panv3.eps}}
  \subfigure[\scriptsize{6.29/0.4270}]{\includegraphics[scale=0.52]{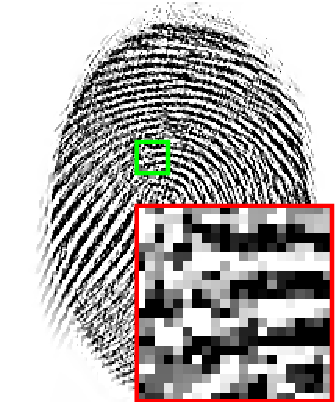}}
  \subfigure[\scriptsize{52.08/0.9999}]{\includegraphics[scale=0.52]{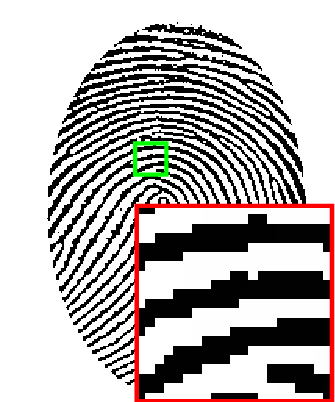}}\\
%  \subfigure[\scriptsize{6.60/0.5805}]{\includegraphics[scale=0.5]{wenv3.eps}}\\
  \subfigure[\scriptsize{Ground truth}]{\includegraphics[scale=0.27]{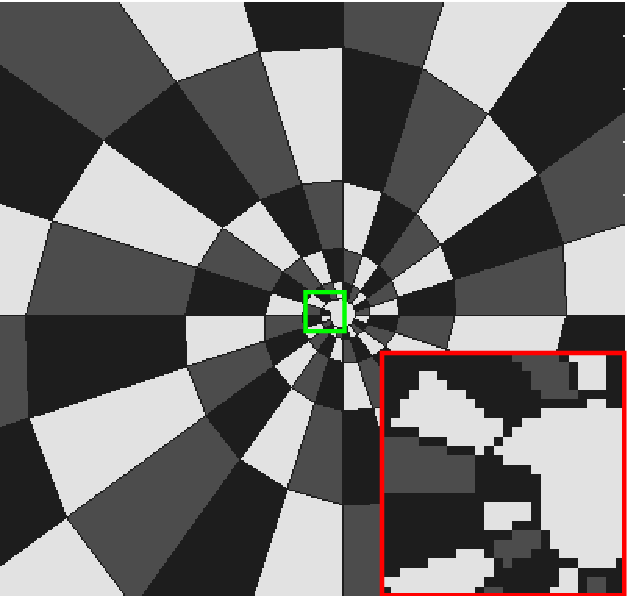}}
  \subfigure[\scriptsize{Blurred image}]{\includegraphics[scale=0.27]{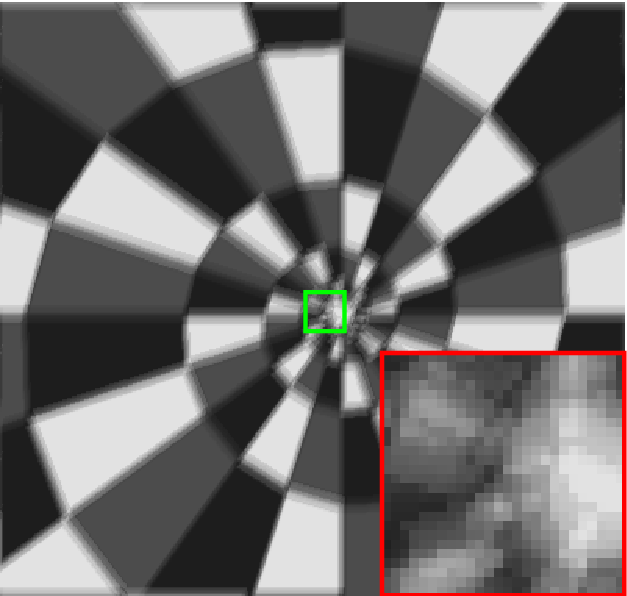}}
  \subfigure[\scriptsize{29.75/0.9682}]{\includegraphics[scale=0.27]{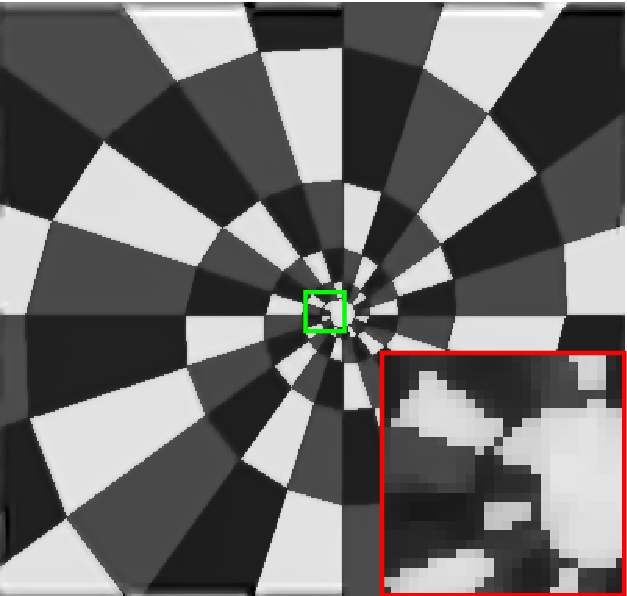}}
  \subfigure[\scriptsize{33.73/0.9922}]{\includegraphics[scale=0.27]{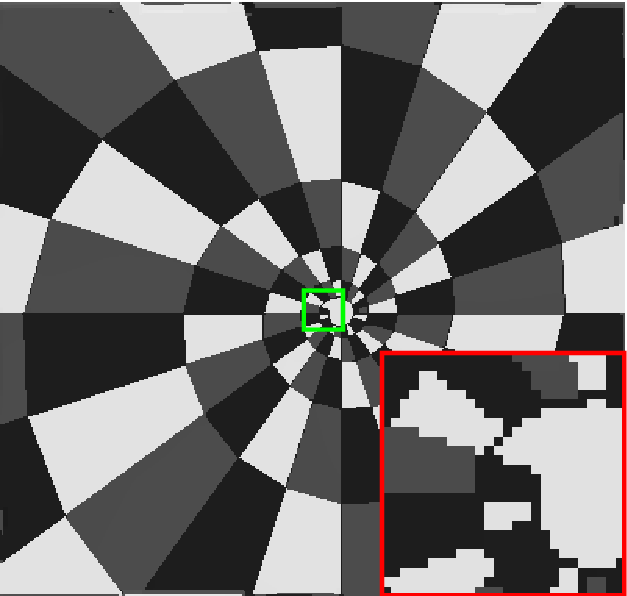}}
  %\subfigure[\scriptsize{81.71/0.9999}]{\includegraphics[scale=0.5]{pano3.eps}}

  %\subfigure[\scriptsize{160.81/0.9999}]{\includegraphics[scale=0.5]{weno3.eps}}
  \caption{Comparisons of \cite{chen2019blind} with and without the pixel intensity constraint. (a) Ground truth, (b) Blurred image, (c) Result of \cite{chen2019blind}, (d) Result of \cite{chen2019blind} with our pixel intensity constraint, (e) Ground truth, (f) Blurred image, (g) Result of \cite{chen2019blind}, (h) Result of \cite{chen2019blind} with our pixel intensity constraint.}
\end{figure}

\section{Conclusion}

In this paper, we introduced a unified framework for blind deblurring of two-valued binary and multi-valued pattern images, by integrating the pixel intensity constraint with an $\ell_0$-based sparsity gradient regularizer. To efficiently address the formulated non-convex and constrained optimization problem, we developed an iterative scheme based on the half-quadratic splitting approach. Experimental results demonstrate that the proposed algorithm consistently outperforms comparable methods in terms of PSNR, SSIM measures, and visual quality.
This preliminary work highlights the importance of leveraging distinct pixel values during the iterative process. Along this research direction, future research could explore and develop sophisticated techniques for more precise estimation of pixel intensities,  further enhancing blind deblurring performance for binary and pattern images.

\end{document}